\documentclass{article} 
\usepackage{iclr2024_conference,times}

\usepackage{amsmath,amsfonts,bm}

\def\eqref#1{equation~\ref{#1}}

\def\1{\bm{1}}

\DeclareMathAlphabet{\mathsfit}{\encodingdefault}{\sfdefault}{m}{sl}
\SetMathAlphabet{\mathsfit}{bold}{\encodingdefault}{\sfdefault}{bx}{n}

\newcommand{\softmax}{\mathrm{softmax}}
\newcommand{\sigmoid}{\sigma}

\usepackage{hyperref}
\usepackage{url}
\usepackage{graphicx}
\usepackage{amsmath}
\usepackage{amssymb}
\usepackage{bm}
\usepackage{caption}
\usepackage{subcaption}
\usepackage{multirow}
\usepackage{bbding}
\usepackage{color}
\usepackage{makecell}
\usepackage{booktabs}
\usepackage{pgfplots}
\usepackage{wrapfig}
\pgfplotsset{compat=1.17}

\title{Image2Sentence based Asymmetric Zero-shot Composed Image Retrieval}

\author{Yongchao Du$^{1}$, Min Wang$^{2*}$, Wengang Zhou$^{1,2}$\thanks{Corresponding authors: Min Wang and Wengang Zhou}   , Shuping Hui$^{1}$, Houqiang Li$^{1,2}$ \\
$^{1}$CAS Key Laboratory of Technology in GIPAS, University of Science and Technology of China \\
$^{2}$Institute of Artificial Intelligence, Hefei Comprehensive National Science Center \\
\texttt{\{ycdu2020,huisp\}@mail.ustc.edu.cn, wangmin@iai.ustc.edu.cn, } \\ \texttt{\{zhwg,lihq\}@ustc.edu.cn} \\
}

\iclrfinalcopy 
\begin{document}

\maketitle

\begin{abstract}
The task of composed image retrieval (CIR) aims to retrieve images based on the query image and the text describing the users' intent. 
Existing methods have made great progress with the advanced large vision-language (VL) model in CIR task, however, they generally suffer from two main issues: lack of labeled triplets for model training and difficulty of deployment on resource-restricted environments when deploying the large vision-language model.
To tackle the above problems, we propose Image2Sentence based Asymmetric zero-shot composed image retrieval (ISA), which takes advantage of the VL model and only relies on unlabeled images for composition learning.
In the framework, we propose a new adaptive token learner that maps an image to a sentence in the word embedding space of VL model. 
The sentence adaptively captures discriminative visual information and is further integrated with the text modifier.
An asymmetric structure is devised for flexible deployment, in which the lightweight model is adopted for the query side while the large VL model is deployed on the gallery side.
The global contrastive distillation and the local alignment regularization are adopted for the alignment between the light model and the VL model for CIR task. 
Our experiments demonstrate that the proposed ISA could better cope with the real retrieval scenarios and further improve retrieval accuracy and efficiency. 

\end{abstract}

\section{Introduction}
Image retrieval is a crucial task for computer vision, and plays a fundamental role in a variety of applications, \textit{e.g.}, product retrieval \cite{dong2022m5product,zhan2021product1m,liu2016deepfashion,zhou2010spatial}, landmark retrieval \cite{radenovic2018fine}, and person re-identification \cite{zheng2016person,gheissari2006person}.
Composed image retrieval (CIR) is presented as a more challenging task that requires accurate understanding of the users' intentions. 
CIR allows multi-modal queries and aims to retrieve images that are both visually similar to the given query image and match with the text modifier, which enables users to conduct more precise and flexible search.
However, to learn the composition ability, it requires labeled triplets of reference image, target image and text modifier \cite{liu2021image,wu2021fashion}, which are time-consuming and expensive to collect.

To eliminate the limitation of data scarcity, zero-shot composed image retrieval (ZSCIR) is proposed in \cite{saito2023pic2word,baldrati2023zero}, which converts the CIR task to text-to-image retrieval by mapping the image into a single word and combining it with text modifier. 
However, they all adopt a symmetric retrieval setting based on CLIP models \cite{radford2021learning}, \textit{i.e.}, using large models to extract both query and database features, and fail to consider the real-life application where the model may be deployed on a resource-limited platform such as mobile device, etc. 
Moreover, the mapping is implemented by an MLP that projects image feature to a single word vector, \textit{i.e.}, representing image as a single word in the word embedding space.
This may be insufficient as a faithful representation for an image, and the heterogeneity of word embedding space and visual feature space may further dampen the discriminative information of image feature after projection, thus constraining its potential.

To address the above issues, we propose Image2Sentence based Asymmetric ZSCIR that utilizes different models to extract query and database features.
Specifically, a lightweight model is adopted for the query side, which is usually user's terminal device with limited computation resources; the large VL model is employed for the gallery side, which is usually deployed on a cloud platform. 
Asymmetric structures often suffer from two core issues: inferior representation capability and misalignment between the lightweight model and the larger model.
To this end, we propose an adaptive token learner to realize the alignment between the lightweight visual model and the large visual language model, which could further improve the representation discrimination ability of the lightweight model.
Specifically, the lightweight model extracts the visual feature map of an input image, then the adaptive token learner extracts visual tokens from visual feature map, and maps them to a more informative sentence in the word embedding space. 
By adaptively focusing on more distinctive visual patterns, our adaptive token learner serves as a semantic filter that automatically filters out trivial background noise and preserves discriminative visual information. 
The mapped sentence is learned to cope with the visual feature from the large visual encoder under the global contrastive distillation (GCD), as a constraint to build up semantic alignment between the large model and the lightweight model.
Therefore during inference, the mapped sentence is directly concatenated with the modifier text as a caption to search for images.   
To better learn the image2sentence mapping and retain more visual information, we further leverage the local alignment regularization (LAR) to help the mapped sentence capture more distinctive visual concepts, and to be closer to the real caption that faithfully describes the image. 
Our asymmetric framework allows for more flexible deployment while also enhancing performance compared with symmetric setting, and outperforms the state-of-the-art methods on three benchmarks.

\vspace{-0.5em}
\section{Related Work}
\subsection{Composed Image Retrieval}
\vspace{-0.5em}
CIR is the task of retrieving images that correspond to the reference image with a modifier text, and has been investigated in several domains, \textit{e.g.}, natural scene \cite{baldrati2023zero,liu2021image}, fashion domain \cite{dodds2022training,lee2021cosmo,wu2021fashion}, and synthetic domain \cite{gu2023compodiff,vo2019composing}.
Supervised learning serves as a stepping stone in the progression of CIR, and is first introduced in \cite{vo2019composing}, in which meticulously annotated (reference image, text modifier, target image) triplets are adopted for training.
Supervised methods mainly follow the pipeline that merges the features of reference image and text modifier to align with the target image feature, as a result, their innovations generally revolve around jointly capturing the information from both visual and textual modality \cite{vo2019composing,lee2021cosmo,chen2020image}. 
Moreover, some methods further improve the performance by adopting multimodal foundation model \cite{baldrati2021conditioned,baldrati2022conditioned,baldrati2022effective,baldrati2023composed,levy2023data} or incorporating pretraining-finetuning strategy \cite{dodds2022training,han2022fashionvil,han2023fame}. 

However, the above supervised methods heavily rely on the triplet training dataset, which involves great difficulty in collecting and labeling in large scale.
A new task of zero-shot composed image retrieval (ZSCIR) \cite{saito2023pic2word} is proposed to liberate the model learning from meticulous triplet-labeled data, and relies only on image-text pairs.
\cite{saito2023pic2word} propose to use MLP to convert CLIP visual feature to a single word in the word embedding space, and adopt a cycle contrastive loss for training; \cite{baldrati2023zero} propose a two-stage framework including text inversion and distillation to convert image into text domain.  
However, the above methods rely on symmetric retrieval, in which heavy models are utilized to extract features for both query and database, and are difficult to deploy in practical retrieval application scenarios.
We propose a ZSCIR framework, which enables flexible deployment on resource-limited scenarios by adopting the lightweight model. 
Different from previous works, instead of converting image to a single word, we propose the adaptive token learner to filter out more discriminative visual information, which is further mapped to a more informative sentence.
Apart from global contrastive loss, a dense local alignment regularization is applied on mapped sentences to preserve the visual information during conversion. 
Our work shows its superiority in both retrieval accuracy and deployment flexibility.

\subsection{Asymmetric Retrieval}
In recent times, the adoption of expansive deep models has greatly improved the multi-modal comprehension ability ~\cite{jia2021scaling,li2021align,li2022blip,li2023blip,radford2021learning}. 
However, large pre-trained models usually consume more resources in computing and memory, which is not desirable for some resource-constrained scenarios such as image retrieval.
To tackle this inconvenience, recently many works \cite{budnik2021asymmetric,wu2022general,duggal2021compatibility,wu2022contextual} have been proposed to research into asymmetric image retrieval, in which the query (user) side adopts a lightweight model such as  \cite{howard2017mobilenets,howard2019searching,liu2023efficientvit,mehta2021mobilevit,sandler2018mobilenetv2,tan2019efficientnet} while the gallery (server) side takes a large model. 
To our knowledge, our proposed framework is the first work to adopt asymmetric retrieval in CIR task, and could even outperform its symmetric counterpart.
\vspace{-1.0em}

\section{Method}
\vspace{-1.0em}
\begin{figure*}[t]
	\begin{center}
		\includegraphics[width=0.96\textwidth]{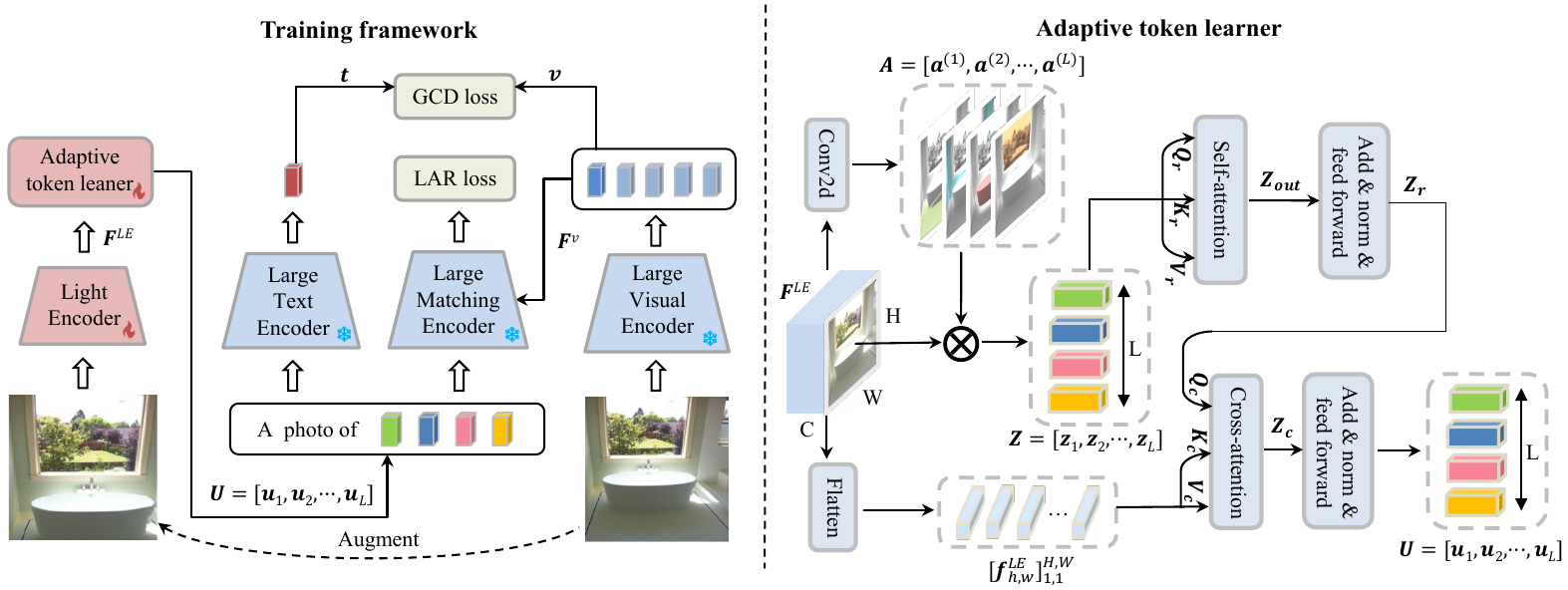} 
	\end{center}
	\vspace{-1.0em}
	\caption{
		\textbf{Left}: our training framework. Large visual encoder, large text encoder and matching encoder of pre-trained foundation model are frozen during training.
		The framework is trained with global contrastive distillation (GCD) and local alignment regularization (LAR).
		\textbf{Right}: adaptive token learner.
		Adaptive token learner learns to adaptively select salient visual patterns and map them to a sentence in the word embedding space of text encoder with attention mechanism. 
	} 
	\label{framework}
	\vspace{-1.3em}
\end{figure*}

The overview of our framework is shown in Figure \ref{framework}. 
Our framework integrates a pre-trained vision-language (VL) model, comprising a large visual encoder and a large text encoder. These encoders are specifically designed to align visual and textual modalities within a unified feature space. Additionally, a matching encoder is incorporated to enhance image-text matching capabilities, which ensures fine-grained alignment and interpretation between different modalities.
We freeze the parameter of VL model to maintain the alignment between visual and textual modality during training. 
Given an image $\boldsymbol{I}$, a lightweight visual encoder is applied to extract visual feature map $\boldsymbol{F}^{LE}$ of augmented $\boldsymbol{I}$, and a new adaptive token learner is utilized to map the visual feature map to a series of sentence tokens $\boldsymbol{U}$. 
$\boldsymbol{U}$ is then concatenated with a prompt such as ``a photo of" to form a full sentence, which is fed into the large text encoder to obtain textual feature $\boldsymbol{t}$.
Large visual encoder is utilized to extract the image feature $\boldsymbol{v}$ and feature map $\boldsymbol{F}^{v}$ of $\boldsymbol{I}$.
Global contrastive distillation is applied to align $\boldsymbol{v}$ and $\boldsymbol{t}$ while local alignment regularization is applied on $\boldsymbol{U}$ and $\boldsymbol{F^{v}}$.

\subsection{Light Visual Encoder and Adaptive Token Learner}
\vspace{-0.5em}
As shown in Figure \ref{framework}, we apply data augmentation to input image $\boldsymbol{I}$ and feed it into the light visual encoder, to extract visual feature map $\boldsymbol{F}^{LE} \in \mathbb{R}^{H \times W \times C}$.
It is a common understanding that light models are less discriminative compared with the large models. 
Therefore, we introduce an innovative adaptive token learner. Instead of representing an image as a single word, our method conceptualizes the image as a sentence, which significantly enhances visual discrimination within the word embedding space. Each image element is treated as an individual narrative component, allowing for a richer and more detailed representation in the embedding space.
Specifically, we map each pixel $\boldsymbol{f}^{LE}_{h,w}$ of $\boldsymbol{F}^{LE}$ to one of $L$ semantic groups using convolutions, and generate $L$ attention maps $\boldsymbol{A} = \{\boldsymbol{a}^{(1)}, \cdots, \boldsymbol{a}^{(L)}\}$ conditioned on the $\boldsymbol{F}^{LE}$.
The spatial attention maps are computed as
\vspace{-0.3em}
\begin{equation}
	\boldsymbol{a}^{(l)}_{h, w} = \frac{\exp(\boldsymbol{w}^{(l)} \cdot \boldsymbol{f}^{LE}_{h, w})}{\sum_{i=1}^{L} \exp(\boldsymbol{w}^{(l)} \cdot \boldsymbol{f}^{LE}_{h,w})}, \label{spatial attention map}
	\vspace{-0.5em}
\end{equation}
where $\boldsymbol{W} = [\boldsymbol{w}^{(1)}, \cdots, \boldsymbol{w}^{(L)}] \in \mathbb{R}^{L \times C}$ are the parameters of the $L$ convolutions.
Spatial attention is applied to adaptively discover discriminative visual patterns, and obtain $L$ visual tokens $\boldsymbol{Z} = [\boldsymbol{z}^{(1)}, \cdots, \boldsymbol{z}^{(L)}]$
\vspace{-1.7em}
\begin{equation}
	\boldsymbol{z}^{(l)} = \frac{1}{\rho(\boldsymbol{a}^{(l)})} \sum_{h=1, w=1}^{H, W} \boldsymbol{a}^{(l)}_{h,w} \cdot \boldsymbol{f}^{LE}_{h,w},	\quad
	\rho(\boldsymbol{a}^{(l)}) =  \sum_{h=1, w=1}^{H, W} \boldsymbol{a}^{(l)}_{h,w}.
	\label{visual token}
\end{equation}
Since each visual token is learned separately without considering their mutual relationship, self-attention is utilized to model the interaction among visual tokens to obtain the refined tokens $\boldsymbol{Z}_{r}$
\begin{equation}
	\begin{aligned}
		\boldsymbol{Z}_{out} =& \boldsymbol{Z} + \softmax_{L}((\boldsymbol{Z} \boldsymbol{K}_{r})(\boldsymbol{Z} \boldsymbol{Q}_{r})^{T})\boldsymbol{Z}\boldsymbol{V}_{r}, \\
		& \boldsymbol{Z}_{r} = \boldsymbol{Z}_{out} + \sigmoid(\boldsymbol{Z}_{out}\boldsymbol{W}_{r1})\boldsymbol{W}_{r2}, 
	\end{aligned}
	\label{self-attention}
\end{equation}
where $\boldsymbol{Q}_{r}, \boldsymbol{K}_{r}, \boldsymbol{V}_{r}$ are the parameters of query, key and value, and $\boldsymbol{W}_{r1}, \boldsymbol{W}_{r2}$ are the parameters of feed-forward network with $\sigmoid(\cdot)$ activation. 
To obtain the $L$ sentence tokens that involve filtered visual semantics, we adopt the cross-attention to adaptively select the salient visual patterns from the feature map $\boldsymbol{F}^{LE}$. 
$\boldsymbol{F}^{LE}$ is first flattened as a sequence $\boldsymbol{F}^{LE'}=[\boldsymbol{f}^{LE}_{1,1}, \cdots, \boldsymbol{f}^{LE}_{H,W}] \in \mathbb{R}^{HW \times C}$, and the sentence tokens are calculated as 
\begin{equation}
	\begin{aligned}
		\boldsymbol{Z}_{c} = \boldsymbol{Z}_{r} + & \softmax_{L}((\boldsymbol{F}^{LE'} \boldsymbol{K}_{c})(\boldsymbol{Z}_{r} \boldsymbol{Q}_{c})^{T})\boldsymbol{F}^{LE'}\boldsymbol{V}_{c}, \\
		& \boldsymbol{U} = [\boldsymbol{Z}_{c} + \sigmoid(\boldsymbol{Z}_{c}\boldsymbol{W}_{c1})\boldsymbol{W}_{c2}] \boldsymbol{W}_{p}, 
	\end{aligned}
	\label{cross-attention}
\end{equation}
where $\boldsymbol{Q}_{c}, \boldsymbol{K}_{c}, \boldsymbol{V}_{c}$ are the parameters of query, key and value, and $\boldsymbol{W}_{c1}, \boldsymbol{W}_{c2}$ are the parameters of feed-forward network with $\sigmoid(\cdot)$ activation.
$\boldsymbol{W}_{p} \in \mathbb{R}^{C \times d_{w}}$ are the parameters of fc layer to project the output from feed-forward network to $L$ sentence tokens $\boldsymbol{U} = [\boldsymbol{u}^{(1)}, \cdots, \boldsymbol{u}^{(L)}] \in \mathbb{R}^{L \times d_{w}}$, where $d_{w}$ is the dimension of word embedding space.
In sum, the adaptive token learner incorporates the spatial attention mechanism to filter out noisy information by selecting more prominent visual patterns, and map them to the word embedding space as a sentence.

\subsection{Learning with Pre-trained Foundation Model} 
Given sentence tokens $\boldsymbol{U}$, we concatenate them with the predefined prompt to form a full sentence, and feed it into the large text encoder to obtain the textual feature $\boldsymbol{t}$.
During training, $\boldsymbol{t}$ is constrained to be similar to the corresponding image feature $\boldsymbol{v}$ extracted from the large visual encoder within batch $\mathcal{B}$.
The normalized image-to-text and text-to-image similarity are calculated as 
\begin{equation}
	\boldsymbol{p}^{I2T} = \frac{\exp(\boldsymbol{v}_{i}^{T} \boldsymbol{t}_{i} / \tau)}{\sum\limits_{j \in \mathcal{B}} \exp(\boldsymbol{v}_{i}^{T} \boldsymbol{t}_{j} / \tau)}, ~~ \boldsymbol{p}^{T2I} = \frac{\exp(\boldsymbol{t}_{i}^{T} \boldsymbol{v}_{i} / \tau)}{\sum\limits_{j \in \mathcal{B}} \exp(\boldsymbol{t}_{i}^{T} \boldsymbol{v}_{j} / \tau)}, \label{ITC probability}
\end{equation}
where $\tau$ denotes the temperature parameter.
Suppose $\boldsymbol{y}^{I2T}, \boldsymbol{y}^{T2I}$ denote the ground-truth one-hot similarity, the global contrastive distillation is computed as the cross-entropy $H$ between $\boldsymbol{p}$ and $\boldsymbol{y}$
\begin{equation}
	\mathcal{L}_{GCD} = \frac{1}{2} \mathbb{E}_{\mathcal{B}}[H(\boldsymbol{p}^{I2T}, \boldsymbol{y}^{I2T}) + H(\boldsymbol{p}^{T2I}, \boldsymbol{y}^{T2I})]. \label{ITC loss}
\end{equation}
To further enhance the semantic alignment of sentence tokens with visual information, local alignment regularization is applied to harness sentence tokens to be more like the caption that faithfully describes the image content in the word embedding space.
Suppose that the word embedding space is shared across the text encoder and matching encoder, therefore the local alignment regularization will help to generate better sentence tokens, and improve the global semantic alignment between $\boldsymbol{t}$ and $\boldsymbol{v}$.
Specifically, we feed sentence tokens together with visual feature map $\boldsymbol{F}^{v}$ into the matching encoder to obtain the probability $\boldsymbol{p}^{LAR}$.
The local alignment regularization is calculated as
\begin{equation}
	\mathcal{L}_{LAR} = \mathbb{E}_{\mathcal{B}}[H(\boldsymbol{p}^{LAR}, \boldsymbol{y}^{LAR})], \label{itm loss}
\end{equation}
where $\boldsymbol{y}^{LAR}$ denotes the binary one-hot vector that represents the paired label. 
Therefore, the final loss could be computed as
\begin{equation}
	\mathcal{L} = \mathcal{L}_{GCD} + \mathcal{L}_{LAR}. \label{total loss}
\end{equation}
With the knowledge from the pre-trained foundation model, our framework could acquire composition ability by training on unlabeled images.

\begin{figure*}[htb]
	\begin{center}
		\includegraphics[width=0.95\textwidth]{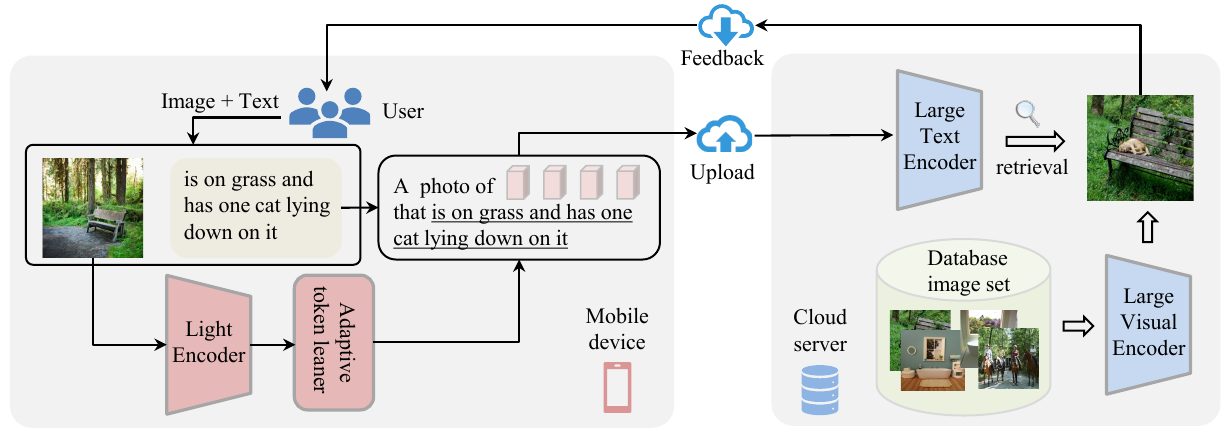} 
	\end{center}
	\vspace{-1.0em}
	\caption{
		The inference workflow of our framework.
		The light encoder and the adaptive token learner are deployed on mobile device, to extract sentence tokens from the query image and concatenate them with the prompt and text modifier as the composed query.  
		Pre-trained foundation model is deployed on the cloud server, and the large text encoder receives the uploaded composed query to extract the text feature for retrieval, and the large visual encoder is used to extract features of database images. 
		The retrieval results are returned to the user side. 
	} 
	\label{inference}
	\vspace{-1.0em}
\end{figure*}

\subsection{Inference}
In practical application scenarios, the memory and computing resources should be included in our consideration.
Therefore, as shown in Figure \ref{inference}, the lightweight visual encoder and the adaptive token learner are deployed on mobile devices, while the large visual and text encoder are deployed on the server side. 
During the inference phase, the user inputs a query image along with a text modifier. The system then converts this query image into sentence tokens, a process that encapsulates the visual content into a textual format. This conversion is pivotal for the subsequent retrieval process, as it enables the system to interpret and process the visual data in a linguistically contextualized manner.
Specifically, we concatenate the sentence tokens and pre-defined prompts together with text modifier as the composed query: [prompt] $\boldsymbol{u}_{1}, \boldsymbol{u}_{2}, \cdots, \boldsymbol{u}_{L}$ that [text modifier].
It is uploaded to the cloud server to retrieve the images from the database. 
The large text encoder is utilized to extract the composed query feature during online inference, and the large visual encoder is only utilized to extract image features from the database offline.

\section{Experiments}
In this section, we conduct a series of experiments to evaluate our ISA on the task of zero-shot composed image retrieval. 
Three datasets are utilized to verify the effectiveness of our method, including CIRR \cite{liu2021image}, FashionIQ \cite{wu2021fashion} and CIRCO \cite{baldrati2023zero}. 

\subsection{Implementation Detail}
We adopt the BLIP model \cite{li2022blip} pre-trained on 129M image-text pairs as the multi-modal foundation model.
CC3M \cite{sharma2018conceptual} is adopted as the training set in our framework.
\begin{wraptable}{r}{0.6\textwidth}
	\centering
	\renewcommand\arraystretch{1.1}
	\resizebox{0.55\textwidth}{0.06\textheight}{
		\begin{tabular}{cccccc}
			\toprule[2.0pt]
			\multirow{2}{*}{\textbf{Query model}} & \multirow{2}{*}{\textbf{Gallery model}} & \multicolumn{2}{c}{\textbf{FLOPS(G)}} & \multicolumn{2}{c}{\textbf{PARAM(M)}} \\
			\cline{3-6}
			& & ABS & \% & ABS & \% \\ 
			\hline
			BLIP VE & \makecell[c]{BLIP VE} & 16.86 & 100.0 & 87.0 & 100.0 \\ 
			\hline
			EfficientNet B0 & \multirow{5}{*}{\makecell[c]{BLIP VE}} & 0.43 & 2.6 & 5.3 & 6.1 \\
			EfficientNet B2 & & 0.72 & 4.3 & 8.5 & 9.7 \\
			EfficientViT M2 & & 0.21 & 1.2 & 5.0 & 5.7 \\
			MobileNet V2 & & 0.35 & 2.1 & 3.5 & 4.0 \\ 
			MobileViT V2 & & 1.42 & 8.4 & 5.5 & 6.3 \\
			\bottomrule[2.0pt]
		\end{tabular}
	}
	\caption{
		FLOPS and parameters of different light models with adaptive token learner in our work, absolute ~(ABS) and relative (\%) to the BLIP visual encoder (BLIP VE). 
	} 
	\label{tab:model FLOPS and parameters}
	\vspace{-1.0em}
\end{wraptable}
Since BLIP visual encoder is frozen during training, image feature $\boldsymbol{v}$ and visual feature map $\boldsymbol{F}^{v}$ could be extracted and stored in advance. 
For the adaptive token learner, the token length $L$ is set as 6, and two hidden dimensions of feed-forward are set as 256 and 512, respectively.
AdamW optimizer with 3e-4 learning rate is adopted, and the framework is trained for 20 epochs with 5 epochs of linear warm-up and 15 epochs of cosine annealing.
The proposed method is implemented on open source Pytorch framework on a server with 4 NVIDIA GeForce 	RTX 3090 GPU, with the batch size as 320.

We first compare with three simple baselines which directly extract the visual and text features using large foundation model: (i) Image-only, using only visual features for both query and database; (ii) Text-only, using textual features for query and visual features for database; (iii) Image + Text, simply merging the visual and textual features with addition for query, while using the visual features for database. 
For the state-of-the-art ZSCIR methods, Pic2word \cite{saito2023pic2word}and SEARLE \cite{baldrati2023zero} are included.
Since Pic2word and SEARLE are all based on ViT-L/14 CLIP model, we re-implement their methods on BLIP with the open-source codes for fair comparison. 
Two lightweight models are mainly used as the lightweight visual encoder for our ISA: EfficientNet B2 \cite{tan2019efficientnet} and EfficientViT M2 \cite{liu2023efficientvit}, as representatives of two main categories of deep architectures (CNN and transformer). 
Since CLIP models do not include the pre-trained matching encoder that share the word embedding space with the text encoder, we only report our results based on BLIP.
We also compare ISA with a symmetric baseline, \textit{i.e.}, frozen BLIP visual encoder is adopted to obtain the visual feature map, followed by the same adaptive token learner to obtain the sentence tokens. 

\vspace{-0.5em}
\begin{table}[htb]
	\centering
	\renewcommand\arraystretch{1.1}
	\resizebox{0.92\textwidth}{0.11\textheight}{
		\begin{tabular}{cccccccc}
			\toprule[2.0pt]
			\multirow{2}{*}{\textbf{Gallery model}} &
			\multirow{2}{*}{\textbf{Query model}} & \multirow{2}{*}{\textbf{Method}} & \multicolumn{4}{c}{\textbf{Recall@K}} &
			\multirow{2}{*}{\textbf{Avg}} \\
			\cline{4-7}
			& & & R@1 & R@5 & R@10 & R@50 & \\ 
			\hline
			\multirow{5}{*}{\makecell[c]{CLIP \\ ViT-L/14}} & CLIP VE & Image-only & 7.40 & 23.60 & 34.00 & 57.40 & 30.60 \\
			& - & Text-only & 20.90 & 44.80 & 56.70 & 79.10 & 50.36 \\
			& CLIP VE & Image + Text & 12.40 & 36.20 & 49.10 & 78.20 & 43.97 \\
			& CLIP VE & Pic2Word \cite{saito2023pic2word} & 23.90 & 51.70 & 65.30 & 87.80 & 57.17 \\
			& CLIP VE & SEARLE \cite{baldrati2023zero} & 24.22 & 52.41 & 66.29 & 88.63 & 57.88 \\
			\midrule[1.5pt]
			\multirow{8}{*}{BLIP} & BLIP VE & Image-only & 7.52 & 21.83 & 31.21 & 52.94 & 28.37 \\
			& - & Text-only & 26.48 & 50.46 & 61.22 & 82.26 & 55.10 \\
			& BLIP VE & Image + Text & 8.19 & 24.19 & 34.41 & 57.81 & 31.15 \\
			& BLIP VE & Pic2word \cite{saito2023pic2word} & 26.70 & 53.16 & 64.10 & 84.36 & 57.08 \\
			& BLIP VE & SEARLE \cite{baldrati2023zero} & 29.27 & 54.86 & 66.57 & 86.16 & 59.21 \\
			& BLIP VE & \textbf{Ours(Sym)} & 29.68 & 58.72 & 70.79 & 90.33 & 62.38 \\
			& EfficientNet B2 & \textbf{Ours(Asy)} & \textbf{30.84} & \textbf{61.06} & \textbf{73.57} & \textbf{92.43} & \textbf{64.48} \\
			& EfficientViT M2 & \textbf{Ours(Asy)} & 29.63 & 58.99 & 71.37 & 91.47 & 62.87 \\
			\bottomrule[2.0pt]
		\end{tabular}
	}
	\caption{
		Results of zero-shot composed image retrieval on CIRR test set.
		The best result for each column is shown in bold fonts. 
		Sym: use BLIP visual models to both extract features for query and database images.
		Asy: use different visual models to extract features for query and database images.
	} 
	\label{tab:SOTA ZSCIR CIRR}
	\vspace{-1.0em}
\end{table}

\vspace{-0.3em}
\begin{table}[htb]
	\centering
	\renewcommand\arraystretch{1.1}
	\resizebox{0.95\textwidth}{0.1\textheight}{
		\begin{tabular}{cccccccc}
			\toprule[2.0pt]
			\multirow{2}{*}{\textbf{Gallery model}} &
			\multirow{2}{*}{\textbf{Query model}} & \multirow{2}{*}{\textbf{Method}} & \multicolumn{4}{c}{\textbf{mAP@K}} &
			\multirow{2}{*}{\textbf{Avg}} \\
			\cline{4-7}
			& & & mAP@5 & mAP@10 & mAP@25 & mAP@50 & \\ 
			\hline
			\multirow{5}{*}{\makecell[c]{CLIP \\ ViT-L/14}} & CLIP VE & Image-only & 1.69 & 2.18 & 2.78 & 3.22 & 2.47 \\
			& - & Text-only & 2.67 & 3.00 & 3.38 & 3.69 & 3.19 \\
			& CLIP VE & Image + Text & 3.60 & 4.40 & 5.59 & 6.21 & 4.95 \\
			& CLIP VE & Pic2Word \cite{saito2023pic2word} & 8.43 & 9.11 & 10.15 & 10.73 & 9.61 \\
			& CLIP VE & SEARLE \cite{baldrati2023zero} & 10.18 & 11.03 & 12.72 & 13.67 & 11.90 \\
			\midrule[1.5pt]
			\multirow{8}{*}{BLIP} & BLIP VE & Image-only & 1.28 & 1.48 & 1.88 & 2.13 & 1.69 \\
			& - & Text-only & 4.01 & 4.37 & 4.79 & 5.13 & 4.58 \\
			& BLIP VE & Image + Text & 1.39 & 1.72 & 2.22 & 2.51 & 1.96 \\
			& BLIP VE & Pic2word \cite{saito2023pic2word} & 8.69 & 9.36 & 10.40 & 10.99 & 9.86 \\
			& BLIP VE & SEARLE \cite{baldrati2023zero} & 10.65 & 11.34 & 12.40 & 13.02 & 11.85 \\
			& BLIP VE & \textbf{Ours(Sym)} & 9.67 & 10.32 & 11.26 & 11.61 & 10.72 \\
			& EfficientNet B2 & \textbf{Ours(Asy)} & \textbf{11.33} & \textbf{12.25} & \textbf{13.42} & \textbf{13.97} & \textbf{12.74} \\
			& EfficientViT M2 & \textbf{Ours(Asy)} & 9.82 & 10.50 & 11.61 & 12.09 & 11.01 \\
			\bottomrule[2.0pt]
		\end{tabular}
	}
	\caption{
		Results of zero-shot composed image retrieval on CIRCO test set.
		The best result for each column is shown in bold fonts. 
		Sym: use BLIP visual models to both extract features for query and database images.
		Asy: use different visual models to extract features for query and database images.
	} 
	\label{tab:SOTA ZSCIR CIRCO}
	\vspace{-1.0em}
\end{table}

\vspace{-0.3em}
\begin{table}[htb]
	\centering
	\renewcommand\arraystretch{1.1}
	\resizebox{0.98\textwidth}{0.1\textheight}{
		\begin{tabular}{ccccccccccc}
			\toprule[2pt]
			\multirow{2}{*}{\textbf{Gallery model}} &
			\multirow{2}{*}{\textbf{Query model}} & \multirow{2}{*}{\textbf{Method}} & \multicolumn{2}{c}{\textbf{Dress}} & \multicolumn{2}{c}{\textbf{Shirt}} &
			\multicolumn{2}{c}{\textbf{Top\&tee}} &
			\multicolumn{2}{c}{\textbf{Average}} \\
			\cline{4-11}
			& & & R@10 & R@50 & R@10 & R@50 & R@10 & R@50 & R@10 & R@50 \\ 
			\hline
			\multirow{5}{*}{\makecell[c]{CLIP \\ ViT-L/14}} & CLIP VE & Image-only & 5.40 & 13.90 & 9.90 & 20.80 & 8.30 & 17.70 & 7.90 & 17.50 \\
			& - & Text-only & 13.60 & 29.70 & 18.90 & 31.80 & 19.30 & 37.00 & 17.30 & 32.90 \\
			& CLIP VE & Image + Text & 16.30 & 33.60 & 21.00 & 34.50 & 22.20 & 39.00 & 19.80 & 35.70 \\
			& CLIP VE & Pic2Word \cite{saito2023pic2word} & 20.00 & 40.20 & 26.20 & 43.60 & 27.90 & 47.40 & 24.70 & 43.70 \\
			& CLIP VE & SEARLE \cite{baldrati2023zero} & 20.32 & 43.18 & 27.43 & 45.68 & 29.32 & 50.17 & 25.69 & 46.34 \\
			\midrule[1.5pt]
			\multirow{8}{*}{BLIP} & BLIP VE & Image-only & 4.07 & 11.35 & 7.07 & 16.39 & 6.88 & 13.67 & 6.01 & 13.67 \\
			& - & Text-only & 18.10 & 35.05 & 23.45 & 39.21 & 26.06 & 44.93 & 22.54 & 39.73 \\
			& BLIP VE & Image + Text & 5.16 & 13.58 & 8.15 & 18.89 & 7.80 & 16.62 & 7.03 & 16.36 \\
			& BLIP VE & Pic2word \cite{saito2023pic2word} & 21.35 & 42.68 & 27.51 & 46.01 & 29.12 & 49.33 & 25.99 & 46.00 \\
			& BLIP VE & SEARLE \cite{baldrati2023zero} & 22.11 & 41.79 & 29.72 & 48.53 & 31.03 & 52.37 & 27.62 & 47.56 \\
			& BLIP VE & \textbf{Ours(Sym)} & 24.69 & 43.88 & \textbf{30.79} & \textbf{50.05} & \textbf{33.91} & 53.65 & \textbf{29.79} & 49.19 \\
			& EfficientNet B2 & \textbf{Ours(Asy)} & 25.33 & \textbf{46.26} & 30.03 & 48.58 & 33.45 & 53.80 & 29.60 & \textbf{49.54} \\ 
			& EfficientViT M2 & \textbf{Ours(Asy)} & \textbf{25.48} & 45.51 & 29.64 & 48.68 & 32.94 & \textbf{54.31} & 29.35 & 49.50 \\ 
			\bottomrule[2pt]
		\end{tabular}
	}
	\caption{
		Results of zero-shot composed image retrieval on FashionIQ validation set.
		The best result for each column is shown in bold fonts. 
		Sym: use BLIP visual model to both extract features for query and database.
		Asy: use different visual models to extract features for query and database.
	} 
	\label{tab:SOTA ZSCIR fashionIQ}
	\vspace{-2.0em}
\end{table}

\subsection{Comparison with the State-of-the-art Methods}
We compare different methods on the test set of CIRR and CIRCO while on the validation set of FashionIQ, and the final results are computed as the average of 3 runs on different random seeds for training. 
As shown in Table \ref{tab:model FLOPS and parameters}, the lightweight models are more efficient in parameters and computational complexity compared with BLIP visual encoder. 
The retrieval performance on three benchmark datasets is reported in Table \ref{tab:SOTA ZSCIR CIRR}, \ref{tab:SOTA ZSCIR CIRCO}, \ref{tab:SOTA ZSCIR fashionIQ}, respectively. 
Our method could outperform simple baselines (Image-only, Text-only and Image + Text) by a large margin.
From the results of Pic2word and SEARLE, we find that the methods based on BLIP are generally better than those based on CLIP for ZSCIR.
Our EfficientNet B2-based ISA could outperform Pic2word and SEARLE on three benchmark datasets, while EfficientViT M2-based ISA achieves better results on FashionIQ and CIRR and is comparable with SEARLE on CIRCO.
Compared with our symmetric retrieval in Ours (Sym), our ISA could achieve superior performance, which could be attributed to ISA containing more training parameters compared with Ours (Sym) (lightweight model + adaptive token learner v.s. adaptive token learner) during training.
Meanwhile, ISA consumes much less computation resources during online retrieval and is more flexible to deploy for real application scenarios, demonstrating the effectiveness of our framework.

\subsection{Ablation study}
For ablation study, we mainly adopt the validation set of CIRR and CIRCO for evaluation, and the results for FashionIQ are included in the appendix.

\subsubsection{Training with different mapping network and number of tokens}
To verify the effectiveness of our adaptive token learner, we compare it with a variant that adopts MLP as the mapping network, which maps the image feature to a single word following \cite{baldrati2023zero,saito2023pic2word}.
The results are shown in Table \ref{tab:ablation mapping network CIRR/CIRCO}, which demonstrate the consistent superiority for both EfficientNet B2 and EfficientViT M2.
These results prove the effectiveness of our adaptive token learner, which maps the visual feature map to multiple sentence tokens per image, and helps to adaptively select more discriminative visual patterns from the query image to generate more informative sentence tokens.
It is notable that the variant MLP~(single token) could be seen as Pic2word with LAR and outperform Pic2word, thus proving the effectiveness of LAR. 

\vspace{-0.5em}
\begin{table}[htb]
	\centering
	\renewcommand\arraystretch{1.1}
	\resizebox{0.98\textwidth}{0.045\textheight}{
		\begin{tabular}{cccccccccccc}
			\toprule[2.0pt]
			\multirow{2}{*}{\textbf{Query model}} & \multirow{2}{*}{\textbf{Mapping network}} & \multicolumn{4}{c}{\textbf{CIRR}} & \multirow{2}{*}{\textbf{Avg}} &
			\multicolumn{4}{c}{\textbf{CIRCO}} & \multirow{2}{*}{\textbf{Avg}} \\
			\cline{3-6} \cline{8-11}
			& & R@1 & R@5 & R@10 & R@50 &  & mAP@5 & mAP@10 & mAP@25 & mAP@50 &  \\ 
			\hline
			EfficientNet B2 & ATL & \textbf{31.48} & \textbf{63.42} & \textbf{77.27} & \textbf{93.16} & \textbf{66.33} & \textbf{13.19} & \textbf{13.83} & \textbf{15.20} & \textbf{15.85} & \textbf{14.52} \\ 
			EfficientNet B2 & MLP~(single token) & 30.18 & 59.46 & 71.30 & 89.33 & 62.57 & 8.28 & 8.67 & 9.62 & 10.00 & 9.14 \\ 
			\midrule[1.5pt]
			EfficientViT M2 & ATL & 31.26 & 62.31 & 75.70 & 92.38 & 65.41 & 12.24 & 12.75 & 13.80 & 14.27 & 13.27 \\ 
			EfficientViT M2 & MLP~(single token) & 30.32 & 60.25 & 71.49 & 89.57 & 62.90 & 7.19 & 7.73 & 8.47 & 9.09 & 8.12 \\ 
			\bottomrule[2.0pt]
		\end{tabular}
	}
	\caption{
		Results of different mapping networks with BLIP on CIRR and CIRCO validation set.
		ATL denotes the proposed adaptive token learner.
		The best results are shown in bold fonts.
	} 
	\label{tab:ablation mapping network CIRR/CIRCO}
	\vspace{-1.0em}
\end{table}

\begin{figure}[htb] 
	\begin{subfigure}{0.33\textwidth}
		\centering
		\begin{tikzpicture}
			\begin{axis}[
				xlabel={token length of sentence},
				xmin = 0.0, xmax = 11.0,
				ymin = 34.0, ymax = 41.0,
				ylabel={Average scores},
				title={FashionIQ},
				grid=major,
				width=1.1\textwidth,
				height=1.0\textwidth,
				title style={
					font=\fontsize{8}{1}\selectfont
				},
				tick label style={
					font=\fontsize{5}{1}\selectfont
				},
				xlabel style={
					font=\fontsize{7}{1}\selectfont
				},
				ylabel style={
					font=\fontsize{7}{1}\selectfont
				},
			    legend style={
					legend cell align=left, 
					font=\fontsize{4}{1}\selectfont, 
					draw=black, 
					fill=white, 
					inner sep=0.7pt, 
					at={(0.95,0.95)}, 
					anchor=north east 
				},
				legend pos=south east, 
				mark options={
					mark size=1.1pt 
				},
				]
				
				\addplot[
				color=blue,
				solid,
				mark=none,
				line width=0.9pt,
				] coordinates {
					(1, 37.72)
					(2, 39.16)
					(4, 39.55)
					(6, 39.57)
					(8, 39.37)
					(10, 37.29)
				};
				
				\addplot[
				color=blue,
				only marks, 
				forget plot,
				error bars/.cd,
				y dir=both, 
				y explicit, 
				error bar style={
					color=blue!70, 
					line width=0.8pt 
				}, 
				] coordinates {
					(1, 37.72) += (0,0.32) -= (0,0.32)
					(2, 39.16) += (0,0.41) -= (0,0.41)
					(4, 39.55) += (0,0.36) -= (0,0.36)
					(6, 39.57) += (0,0.30) -= (0,0.30)
					(8, 39.37) += (0,0.39) -= (0,0.39)
					(10, 37.29) += (0,0.29) -= (0,0.29)
				};
				
				\addplot[
				color=red,
				dashed,
				mark=none,
				line width=0.9pt,
				] coordinates {
					(1, 37.66)
					(2, 37.59)
					(4, 38.90)
					(6, 39.43)
					(8, 38.46)
					(10, 37.69)
				};
				\addplot[
				color=red,
				solid,
				only marks, 
				forget plot,
				error bars/.cd,
				y dir=both, 
				y explicit, 
				error bar style={
					color=red!70, 
					line width=0.8pt 
				}, 
				] coordinates {
					(1, 37.66) += (0,0.31) -= (0,0.31)
					(2, 37.59) += (0,0.38) -= (0,0.38)
					(4, 38.90) += (0,0.33) -= (0,0.33)
					(6, 39.43) += (0,0.32) -= (0,0.32)
					(8, 38.46) += (0,0.35) -= (0,0.35)
					(10, 37.69) += (0,0.31) -= (0,0.31)
				};
				\addlegendentry{ENT B2}
				\addlegendentry{EVT M2}
			\end{axis}
		\end{tikzpicture}
	\end{subfigure}
	\begin{subfigure}{0.33\textwidth} 
		\centering
		\begin{tikzpicture}
			\begin{axis}[
				xlabel={token length of sentence},
				xmin = 0.0,xmax = 11.0,
				ymin = 61.0, ymax = 68.0,
				ylabel={Average scores},
				title={CIRR},
				grid=major, 
				width=1.1\textwidth, 
				height=1.0\textwidth, 
				title style={
					font=\fontsize{8}{1}\selectfont 
				},
				tick label style={
					font=\fontsize{5}{1}\selectfont 
				},
				xlabel style={
					font=\fontsize{7}{1}\selectfont 
				},
				ylabel style={
					font=\fontsize{7}{1}\selectfont 
				},
				legend style={
					legend cell align=left, 
					font=\fontsize{4}{1}\selectfont, 
					draw=black, 
					fill=white, 
					inner sep=0.7pt, 
					at={(0.95,0.95)}, 
					anchor=north east 
				},
				legend pos=south east,
				mark options={
					mark size=1.1pt 
				},
				]
				
				\addplot[
				color=blue,
				solid,
				mark=none,
				line width=0.9pt,
				] coordinates {
					(1, 62.94)
					(2, 64.06)
					(4, 66.20)
					(6, 66.33)
					(8, 65.08)
					(10, 63.16)
				};
			
				\addplot[
				color=blue,
				only marks, 
				forget plot,
				error bars/.cd,
				y dir=both, 
				y explicit, 
				error bar style={
					color=blue!70, 
					line width=0.8pt 
				}, 
				] coordinates {
					(1, 62.94) += (0,0.42) -= (0,0.42) 
					(2, 64.06) += (0,0.41) -= (0,0.41)
					(4, 66.20) += (0,0.46) -= (0,0.46)
					(6, 66.33) += (0,0.44) -= (0,0.44)
					(8, 65.08) += (0,0.49) -= (0,0.49)
					(10, 63.16) += (0,0.46) -= (0,0.46)
					
				};
				
				\addplot[
				color=red,
				dashed,
				mark=none,
				line width=0.9pt,
				] coordinates {
					(1, 62.11)
					(2, 64.06)
					(4, 66.17)
					(6, 65.41)
					(8, 64.59)
					(10, 62.83)
				};
				
				\addplot[
				color=red,
				solid,
				only marks, 
				forget plot,
				error bars/.cd,
				y dir=both, 
				y explicit, 
				error bar style={
					color=red!70, 
					line width=0.8pt 
				}, 
				] coordinates {
					(1, 62.11) += (0,0.41) -= (0,0.41) 
					(2, 64.06) += (0,0.48) -= (0,0.48)
					(4, 66.17) += (0,0.43) -= (0,0.43)
					(6, 65.41) += (0,0.45) -= (0,0.45)
					(8, 64.59) += (0,0.42) -= (0,0.42)
					(10, 62.83) += (0,0.47) -= (0,0.47)
				};
				\addlegendentry{ENT B2}
				\addlegendentry{EVT M2}
			\end{axis}
		\end{tikzpicture}
	\end{subfigure}
	\begin{subfigure}{0.33\textwidth} 
		\centering
		\begin{tikzpicture}
			\begin{axis}[
				xlabel={token length of sentence},
				xmin = 0.0,xmax = 11.0,
				ymin = 7.0, ymax = 16.0,
				ylabel={Average scores},
				title={CIRCO},
				grid=major,
				width=1.1\textwidth, 
				height=1.0\textwidth, 
				title style={
					font=\fontsize{8}{1}\selectfont
				},
				tick label style={
					font=\fontsize{5}{1}\selectfont
				},
				xlabel style={
					font=\fontsize{7}{1}\selectfont 
				},
				ylabel style={
					font=\fontsize{7}{1}\selectfont 
				},
				legend style={
					legend cell align=left, 
					font=\fontsize{4}{1}\selectfont, 
					draw=black, 
					fill=white,
					inner sep=0.7pt, 
					at={(0.95,0.95)}, 
					anchor=north east 
				},
				legend pos=south east, 
				mark options={
					mark size=1.1pt
				},
				]
				
				\addplot[
				color=blue,
				solid,
				mark=none,
				line width=0.9pt,
				] coordinates {
					(1, 11.73)
					(2, 13.41)
					(4, 14.13)
					(6, 14.52)
					(8, 13.11)
					(10, 12.20)
				};
				
				\addplot[
				color=blue,
				only marks, 
				forget plot,
				error bars/.cd,
				y dir=both, 
				y explicit, 
				error bar style={
					color=blue!70, 
					line width=0.8pt 
				}, 
				] coordinates {
					(1, 11.73) += (0,0.22) -= (0,0.22) 
					(2, 13.41) += (0,0.21) -= (0,0.21)
					(4, 14.13) += (0,0.26) -= (0,0.26)
					(6, 14.52) += (0,0.29) -= (0,0.29)
					(8, 13.11) += (0,0.24) -= (0,0.24)
					(10, 12.20) += (0,0.26) -= (0,0.26)
				};
				
				\addplot[
				color=red,
				dashed,
				mark=none,
				line width=0.9pt,
				] coordinates {
					(1, 8.52)
					(2, 10.49)
					(4, 12.81)
					(6, 13.27)
					(8, 12.21)
					(10, 10.48)
				};
				\addplot[
				color=red,
				solid,
				only marks, 
				forget plot,
				error bars/.cd,
				y dir=both, 
				y explicit, 
				error bar style={
					color=red!70, 
					line width=0.8pt 
				}, 
				] coordinates {
					(1, 8.52) += (0,0.26) -= (0,0.26) 
					(2, 10.49) += (0,0.31) -= (0,0.31)
					(4, 12.81) += (0,0.32) -= (0,0.32)
					(6, 13.27) += (0,0.24) -= (0,0.24)
					(8, 12.21) += (0,0.29) -= (0,0.29)
					(10, 10.48) += (0,0.27) -= (0,0.27)
				};
				\addlegendentry{ENT B2}
				\addlegendentry{EVT M2}
			\end{axis}
		\end{tikzpicture}
	\end{subfigure}
	\vspace{-1.0em}
	\caption{The average and standard deviation of performance scores on three validation sets.}
	\label{different token length}
	\vspace{-0.7em}
\end{figure}
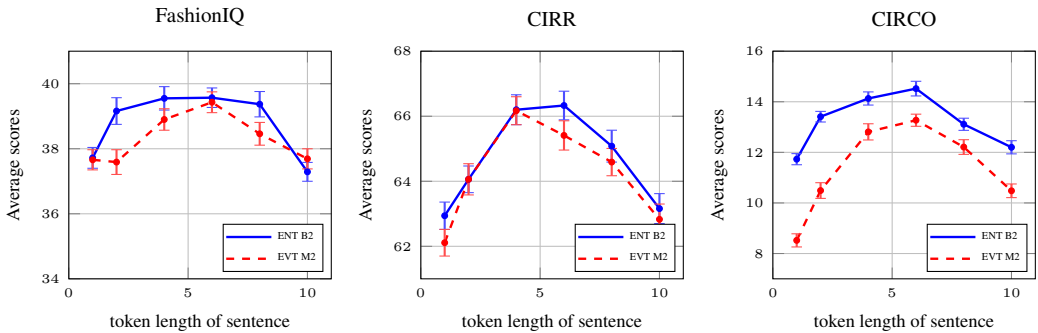

\begin{figure*}[htb]
	\begin{center}
		\includegraphics[width=0.95\textwidth]{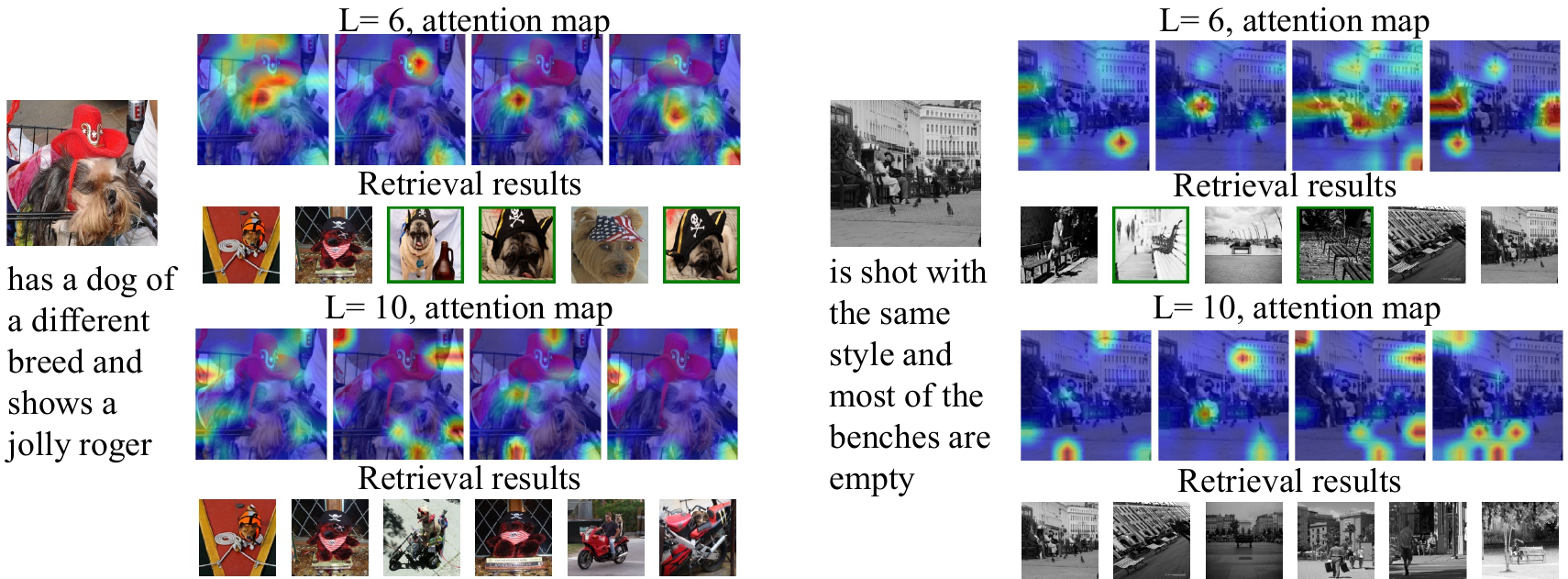} 
	\end{center}
	\vspace{-1.0em}
	\caption{
		Attention map of sentence tokens for different token lengths and retrieval results. 
		Query image and text modifier are on the left side. 
		Ground-truth target images are marked in green boxes.
	} 
	\label{token length attention and retrieval}
	\vspace{-1.0em}
\end{figure*}

For our adaptive token learner, we show the performance of our method with different token lengths $L$ on three datasets in Figure \ref{different token length}.
We average the recall or precision values for each dataset to obtain the overall evaluation criterion.
We find that the average performance improves with the increase of token length for both lightweight visual encoders, however excessive large token length leads to performance degradation. 
We visualize the attention map of adaptive token learner and the retrieval results in Figure \ref{token length attention and retrieval}, and discover that with excessive large token lengths, some sentence tokens may tend to focus on the background or the trivial patterns, which would likely introduce noise to degrade the retrieval performance.

\vspace{-1.0em}

\subsubsection{Train with different Query Models}
In order to investigate the influence of different lightweight encoders, we conduct experiments with different light visual models and pre-training parameters, as shown in Table \ref{tab:ablation light network CIRR/CIRCO}.
We could see that EfficientNet B2 shows the best overall performance in CNN, while EfficientViT M2 is proved to be the best in transformer.  
This could be attributed to the superior network design and larger capacity of parameters. 
Moreover, better initialization weights, \textit{i.e.}, adopting weights pre-trained on ImageNet \cite{deng2009imagenet}, could obtain better performance compared with random initialization. 
This reveals the possibility of adopting different lightweight models to flexibly meet various efficiency and retrieval accuracy requirements, and improve the performance with better pre-trained weights.

\begin{table}[htb]
	\centering
	\renewcommand\arraystretch{1.1}
	\resizebox{0.97\textwidth}{0.085\textheight}{
		\begin{tabular}{ccccccccccc}
			\toprule[2.0pt]
			\multirow{2}{*}{\textbf{Query model}} & \multicolumn{4}{c}{\textbf{CIRR}} & \multirow{2}{*}{\textbf{Avg}} &	\multicolumn{4}{c}{\textbf{CIRCO}} &
			\multirow{2}{*}{\textbf{Avg}} \\
			\cline{2-5} \cline{7-10}
			& R@1 & R@5 & R@10 & R@50 &  & mAP@5 & mAP@10 & mAP@25 & mAP@50 &  \\ 
			\hline
			EfficientNet B0 & \textbf{32.02} & \textbf{64.57} & 76.12 & 92.29 & 66.25 & 11.11 & 12.04 & 13.14 & 13.67 & 12.49 \\ 
			EfficientNet B0* & 27.82 & 58.38 & 71.16 & 90.07 & 61.86 & 9.5 & 9.96 & 10.6 & 11.02 & 10.27 \\ 
			EfficientNet B2 & 31.48 & 63.42 & \textbf{77.27} & \textbf{93.16} & \textbf{66.33} & \textbf{13.19} & \textbf{13.83} & \textbf{15.20} & \textbf{15.85} & \textbf{14.52} \\ 
			EfficientNet B2* & 29.47 & 60.13 & 74.05 & 91.27 & 63.73 & 11.16 & 11.57 & 12.56 & 13.13 & 12.11 \\ 
			EfficientViT M2 & 31.26 & 62.31 & 75.70 & 92.38 & 65.41 & 12.24 & 12.75 & 13.80 & 14.27 & 13.27 \\
			EfficientViT M2* & 28.56 & 59.46 & 71.11 & 89.38 & 62.13 & 10.19 & 10.79 & 11.48 & 11.94 & 11.10 \\ 
			MobileNet V2 & 30.50 & 63.51 & 74.27 & 91.18 & 64.87 & 10.50 & 10.82 & 12.09 & 12.44 & 11.46 \\ 
			MobileNet V2* & 29.47 & 60.14 & 72.10 & 90.07 & 62.95 & 8.55 & 9.16 & 10.04 & 10.39 & 9.54 \\ 
			MobileViT V2 & 27.63 & 58.66 & 70.85 & 88.68 & 61.46 & 9.44 & 9.68 & 10.49 & 10.96 & 10.14 \\
			MobileViT V2* & 27.02 & 57.92 & 69.33 & 88.96 & 60.81 & 8.41 & 8.77 & 9.54 & 9.97 & 9.17 \\
			\bottomrule[2.0pt]
		\end{tabular}
	}
	\caption{
		Results of different query models with BLIP on CIRR/CIRCO validation set.  
		* denotes adopting random initialization for light query model.
		The best results are shown in bold fonts.
	} 
	\label{tab:ablation light network CIRR/CIRCO}
\end{table}

\subsubsection{Symmetric and Asymmetric Retrieval} 
To verify the effectiveness of asymmetric framework, we conduct additional experiments on symmetric setting, which adopts the same model to extract both query and database features.
Specifically, for the symmetric retrieval in the second, 4-th and 6-th rows of Table \ref{tab:ablation symmetrical CIRR/CIRCO}, query and gallery models are both updated during training. 
Therefore, the network should both learn to generate the image feature, and map the image to the sentence tokens in the word embedding space. 
We find that they would drastically degrade the performance. 
This could be attributed to the confusion of multi-task learning for image feature and sentence tokens, since image feature space and word embedding space are heterogeneous semantic spaces.
Moreover, we also carry out additional experiments that adopting BLIP VE as both query model and gallery model, and only query model is updated during training to avoid the multi-task confusion. 
Therefore, BLIP VE is only trained to generate sentence tokens.
As revealed in the result of BLIP VE $\dagger$, there is no breaking down for learning, however it is still outperformed by EfficientNet B2-based ISA.
This could be explained by the fact that BLIP VE is much heavier than light models as shown in Table \ref{tab:model FLOPS and parameters}, and could be trained only with reduced batch size, thus degrading the effectiveness of contrastive and LAR loss.
The above results reveal the importance of pre-trained alignment for different modalities in ZSCIR, but the BLIP visual encoder could be replaced with lightweight model for efficiency.

\vspace{-0.5em}
\begin{table}[htb]
	\centering
	\renewcommand\arraystretch{1.1}
	\resizebox{0.98\textwidth}{0.06\textheight}{
		\begin{tabular}{cccccccccccc}
			\toprule[2.0pt]
			\multirow{2}{*}{\textbf{Query model}} & \multirow{2}{*}{\textbf{Gallery model}} & \multicolumn{4}{c}{\textbf{CIRR}} &
			\multirow{2}{*}{\textbf{Avg}} & \multicolumn{4}{c}{\textbf{CIRCO}} & \multirow{2}{*}{\textbf{Avg}} \\
			\cline{3-6} \cline{8-11}
			& & R@1 & R@5 & R@10 & R@50 & & mAP@5 & mAP@10 & mAP@25 & mAP@50 & \\ 
			\hline
			EfficientNet B2 & BLIP VE & \textbf{31.48} & \textbf{63.42} & \textbf{77.27} & \textbf{93.16} & \textbf{66.33} & \textbf{13.19} & \textbf{13.83} & \textbf{15.20} & \textbf{15.85} & \textbf{14.52} \\ 
			EfficientNet B2 & EfficientNet B2 & 0.61 & 0.74 & 1.02 & 3.97 & 1.59 & 0.00 & 0.02 & 0.08 & 0.14 & 0.06 \\ 
			EfficientViT M2 & BLIP VE & 31.26 & 62.31 & 75.70 & 92.38 & 65.41 & 12.24 & 12.75 & 13.80 & 14.27 & 13.27 \\
			EfficientViT M2 & EfficientViT M2 & 0.29 & 0.37 & 0.79 & 2.68 & 1.03 & 0.00 & 0.01 & 0.04 & 0.11 & 0.04 \\  
			BLIP VE $\dagger$ & BLIP VE & 30.93 & 62.09 & 74.67 & 92.90 & 65.15 & 12.60 & 13.20 & 14.56 & 15.07 & 13.86 \\
			BLIP VE & BLIP VE & 0.49 & 0.60 & 0.83 & 2.96 & 1.22 & 0.02 & 0.03 & 0.08 & 0.15 & 0.07 \\
			\bottomrule[2.0pt]
		\end{tabular}
	}
	\caption{
		Comparison with symmetric retrieval on the validation set of CIRR and CIRCO. 
		BLIP VE denotes the BLIP visual encoder.
		$\dagger$ denotes updating the query BLIP VE while preserving the pre-trained BLIP VE in gallery side during training for maintaining the multi-modality alignment.
		The best results are shown in bold fonts.
	} 
	\label{tab:ablation symmetrical CIRR/CIRCO}
	\vspace{-1.0em}
\end{table}

\subsubsection{Learning with Different Loss Constraints}
In our framework, we mainly apply global contrastive distillation (GCD) and local alignment regularization (LAR) for learning.
We verify the losses by dropping each of them separately. 
To further validate LAR, we compare it with the text regularization, \textit{i.e.}, LLM TR proposed in \cite{baldrati2023zero}. 
Specifically, LLM TR is realized following the pipeline that utilizes a classification model to label each training image and an LLM to do sentence-making.
A cosine loss is applied to constrain the sentence tokens to be similar to the LLM-made sentence as text regularization.
As reported in Table \ref{tab:ablation losses CIRR/CRICO}, it is obvious that both losses are crucial for training, since dropping either of them would lead to consistent worse performance. 
Text regularization from LLM-made sentences would be helpful but only restricted to CIRCO, however it needs extra classification model and LLM model which leads to a much more complex pipeline. 
Our LAR is simple but more effective on three benchmarks.

\vspace{-0.5em}
\begin{table}[htb]
	\centering
	\renewcommand\arraystretch{1.1}
	\resizebox{0.98\textwidth}{0.07\textheight}{
		\begin{tabular}{cccccccccccc}
			\toprule[2.0pt]
			\multirow{2}{*}{\textbf{Query model}} & \multirow{2}{*}{\textbf{Variants}} & \multicolumn{4}{c}{\textbf{CIRR}} & \multirow{2}{*}{\textbf{Avg}} & \multicolumn{4}{c}{\textbf{CIRCO}} & \multirow{2}{*}{\textbf{Avg}} \\
			\cline{3-6} \cline{8-11}
			& & R@1 & R@5 & R@10 & R@50 &  & mAP@5 & mAP@10 & mAP@25 & mAP@50 & \\ 
			\hline
			\multirow{4}{*}{EfficientNet B2} & GCD + LAR & \textbf{31.48} & \textbf{63.42} & \textbf{77.27} & \textbf{93.16} & \textbf{66.33} & \textbf{13.19} & \textbf{13.83} & \textbf{15.20} & \textbf{15.85} & \textbf{14.52} \\ 
			& GCD & 29.86 & 60.18 & 73.90 & 91.13 & 63.77 & 10.46 & 10.98 & 11.88 & 12.41 & 11.43 \\
			& LAR & 27.02 & 56.67 & 68.91 & 88.18 & 60.20 & 7.56 & 7.68 & 8.45 & 8.94 & 8.16 \\
			& GCD + LLM TR & 29.43 & 60.18 & 73.21 & 91.32 & 63.54 & 11.07 & 11.35 & 12.59 & 13.17 & 12.05 \\ 
			\midrule[1.5pt]
			\multirow{4}{*}{EfficientViT M2} & GCD + LAR & \underline{31.26} & \underline{62.31} & \underline{75.70} & \underline{92.38} & \underline{65.41} & \underline{12.24} & \underline{12.75} & \underline{13.80} & \underline{14.27} & \underline{13.27} \\
			& GCD & 26.87 & 55.58 & 69.00 & 89.38 & 60.01 & 9.48 & 9.84 & 10.95 & 11.38 & 10.41 \\ 
			& LAR & 24.35 & 50.49 & 63.26 & 84.00 & 55.53 & 5.75 & 6.00 & 6.98 & 7.26 & 6.50 \\ 
			& GCD + LLM TR & 27.61 & 56.49 & 70.03 & 89.76 & 60.97 & 10.18 & 10.69 & 11.48 & 12.04 & 11.10 \\ 
			\bottomrule[2.0pt]
		\end{tabular}
	}
	\caption{Results of different losses on CIRR and CIRCO validation set.
			GCD: global contrastive distillation; LAR: local alignment regularization; LLM TR: text regularization with the sentences made by LLM.
		The best and second-best results are highlighted in bold and underlined fonts, respectively.} 
	\label{tab:ablation losses CIRR/CRICO}
	\vspace{-1.0em}
\end{table}

\section{Conclusion}
In this paper, we focus on a new but more practical scenarios of asymmetric zero-shot composed image retrieval, which eliminates the requirement of meticulously-labeled training triplets and is flexible for deployment on a resource-constrained platform.
Specifically, lightweight model is adopted for the query side while large visual encoder of foundation model is adopted for the gallery side.
Adaptive token learner is proposed as the semantic filter to automatically select discriminative semantic groups from image and map them to a sentence, maintaining the fidelity of visual semantics within the word embedding space.
By directly concatenating the mapped sentence with text modifier, we are able to compose the multi-modal query to achieve composed retrieval. 
Global contrastive distillation is applied to align the light model with large foundation model, and local alignment regularization is further utilized to help the mapped sentence better align with the real caption that faithfully describes the image.
Extensive experimental results demonstrate that our framework could both achieve better performance and improve retrieval efficiency with deployment flexibility.

\section{Acknowledgement} 
This work was supported by the National Natural Science Foundation of China under Contract 62102128 and 62021001. It was also supported by the GPU cluster built by MCC Lab of Information Science and Technology Institution of USTC.

\bibliography{iclr2024_conference}
\bibliographystyle{iclr2024_conference}

\appendix
\section{Appendix}
In the appendix, we first present detailed descriptions for the three evaluation datasets. 

FashionIQ \cite{wu2021fashion} is a fashion domain dataset, and it includes 77684 fashion images which are divided into three different categories: Dress, Toptee, and Shirt. 
The training set comprises 18000 triplets, and in total 46609 images.
The validation set includes 15537 images and 6017 triplets.
Each query image in the validation set has only one target image.
It also has a test set but without ground-truth labels, thus the test set is not used for evaluation.
 
CIRR \cite{liu2021image} is a dataset of natural domain, and includes 21552 real-life images.
It involves 36554 triplets, and are randomly assigned in 80\% for training, 10\% for validation and 10\% for test.
Each query image in the validation and test set has only one target image.

CIRCO \cite{baldrati2023zero} is collected from MSCOCO and consists of a validation set and a test set, which includes 220 and 800 query images, respectively. 
Each query image has a text modifier and more than one target images. 
The database involves 120k images.

We report the comparison with the state-of-the-art methods on CIRR and CIRCO validation sets in Table \ref{tab:SOTA ZSCIR validation CIRR} and \ref{tab:SOTA ZSCIR validation CIRCO}, respectively.
Accompanied with Table \ref{tab:SOTA ZSCIR CIRR} and \ref{tab:SOTA ZSCIR CIRCO}, our method demonstrates consistent superiority over the simple baselines (Image-only, Text-only and Image + Text) and the state-of-the-art methods of Pic2word and SEARLE. 

\begin{table}[htb]
	\centering
	\renewcommand\arraystretch{1.1}
	\resizebox{0.96\textwidth}{0.14\textheight}{
		\begin{tabular}{cccccccc}
			\toprule[2.0pt]
			\multirow{2}{*}{\textbf{Gallery model}} &
			\multirow{2}{*}{\textbf{Query model}} & \multirow{2}{*}{\textbf{Method}} & \multicolumn{4}{c}{\textbf{Recall@K}} &
			\multirow{2}{*}{\textbf{Average}} \\
			\cline{4-7}
			& & & R@1 & R@5 & R@10 & R@50 & \\ 
			\hline
			\multirow{5}{*}{\makecell[c]{CLIP \\ ViT-L/14}} & \_ & Image-only & 7.06 & 25.52 & 35.64 & 60.18 & 32.10 \\
			& \_ & Text-only & 21.19 & 46.14 & 57.43 & 78.74 & 50.88 \\
			& \_ & Image + Text & 10.79 & 34.37 & 47.05 & 75.27 & 41.87 \\
			& \_ & Pic2Word & 22.54 & 51.69 & 64.76 & 86.19 & 56.30 \\
			& \_ & SEARLE & 25.07 & 55.20 & 68.52 & 90.24 & 59.76 \\
			\midrule[1.5pt]
			\multirow{8}{*}{BLIP} & \_ & Image-only & 6.72 & 23.13 & 32.82 & 55.66 & 29.58 \\
			& \_ & Text-only & 25.38 & 51.02 & 61.76 & 81.68 & 54.96 \\
			& \_ & Image + Text & 7.58 & 25.90 & 36.28 & 61.13 & 32.72 \\
			& \_ & Pic2word & 27.89 & 58.15 & 69.42 & 87.53 & 60.75 \\ 
			& \_ & SEARLE & 28.03 & 60.74 & 71.41 & 88.82 & 62.25 \\
			& \_ & Ours(Sym) & 30.21 & 61.52 & 74.09 & 90.81 & 64.16 \\
			& EfficientNet B2 & Ours(Asy) & 31.48 & 63.42 & \textbf{77.27} & \textbf{93.16} & \textbf{66.33} \\
			& EfficientViT M2 & Ours(Asy) & 31.26 & 62.31 & 75.70 & 92.38 & 65.41 \\
			\bottomrule[2.0pt]	
		\end{tabular}
	}
	\caption{Results of zero-shot composed image retrieval on CIRR validation set.} 
	\label{tab:SOTA ZSCIR validation CIRR}
\end{table}

\begin{table}[htb]
	\centering
	\renewcommand\arraystretch{1.1}
	\resizebox{0.98\textwidth}{0.13\textheight}{
		\begin{tabular}{cccccccc}
			\toprule[2pt] 
			\multirow{2}{*}{\textbf{Gallery model}} &
			\multirow{2}{*}{\textbf{Query model}} & \multirow{2}{*}{\textbf{Method}} & \multicolumn{4}{c}{\textbf{mAP@K}} &
			\multirow{2}{*}{\textbf{Average}} \\
			\cline{4-7}
			& & & mAP@5 & mAP@10 & mAP@25 & mAP@50 & \\ 
			\hline
			\multirow{5}{*}{\makecell[c]{CLIP \\ ViT-L/14}} & \_ & Image-only & 1.85 & 2.24 & 3.11 & 3.57 & 2.69 \\
			& \_ & Text-only & 2.78 & 3.06 & 3.72 & 3.95 & 3.38 \\
			& \_ & Image + Text & 4.71 & 5.74 & 6.88 & 7.44 & 6.19 \\
			& \_ & Pic2Word & 7.50 & 8.31 & 9.45 & 10.11 & 8.84 \\
			& \_ & SEARLE & 11.02 & 12.12 & 13.62 & 14.43 & 12.80 \\
			\midrule[1.5pt] 
			\multirow{8}{*}{BLIP} & \_ & Image-only & 1.86 & 2.18 & 2.71 & 2.97 & 2.43 \\
			& \_ & Text-only & 5.32 & 5.76 & 6.04 & 6.29 & 5.85 \\
			& \_ & Image + Text & 2.04 & 2.63 & 3.18 & 3.49 & 2.84 \\
			& \_ & Pic2Word & 7.91 & 8.27 & 9.05 & 9.56 & 8.70 \\
			& \_ & SEARLE & 11.52 & 11.74 & 12.91 & 13.56 & 12.43 \\
			& \_ & Ours(Sym) & 11.03 & 12.01 & 13.17 & 13.81 & 12.51 \\
			& EfficientNet B2 & Ours(Asy) & \textbf{13.19} & \textbf{13.83} & \textbf{15.20} & \textbf{15.85} & \textbf{14.52} \\
			& EfficientViT M2 & Ours(Asy) & \underline{12.24} & \underline{12.75} & \underline{13.80} & \underline{14.27} & \underline{13.27} \\
			
			\bottomrule[2pt] 
		\end{tabular}
	}
	\caption{Results of zero-shot composed image retrieval on CIRCO validation set.} 
	\label{tab:SOTA ZSCIR validation CIRCO}
	\vspace{-1.0em}
\end{table}

We also report the results of ablation study on FashionIQ dataset. 

Table \ref{tab:ablation mapping network fashionIQ} shows adopting different mapping networks during training, which demonstrates consistent superiority for both EfficientNet B2 and EfficientViT M2.

\begin{table}[htb]
	\centering
	\renewcommand\arraystretch{1.1}
	\resizebox{0.98\textwidth}{0.06\textheight}{
		\begin{tabular}{cccccccccc}
			\toprule[2.0pt]
			\multirow{2}{*}{\textbf{Query model}} & \multirow{2}{*}{\textbf{Mapping network}} & \multicolumn{2}{c}{\textbf{Dress}} & \multicolumn{2}{c}{\textbf{Shirt}} &
			\multicolumn{2}{c}{\textbf{Top\&tee}} &
			\multicolumn{2}{c}{\textbf{Average}} \\
			\cline{3-10}
			& & R@10 & R@50 & R@10 & R@50 & R@10 & R@50 & R@10 & R@50 \\ 
			\hline
			EfficientNet B2 & transformer & 25.33 & \textbf{46.26} & \textbf{30.03} & 48.58 & \textbf{33.45} & 53.80 & \textbf{29.60} & \textbf{49.54} \\
			EfficientNet B2 & MLP~(single token) & 23.65 & 43.83 & 28.46 & 47.74 & 31.16 & 51.96 & 27.76 & 47.84 \\ 
			\midrule[1.5pt]
			EfficientViT M2 & transformer & \textbf{25.48} & 45.51 & 29.64 & \textbf{48.68} & 32.94 & \textbf{54.31} & 29.35 & 49.50 \\  
			EfficientViT M2 & MLP~(single token) & 24.39 & 44.27 & 29.05 & 47.45 & 30.75 & 53.19 & 28.06 & 48.30 \\ 
			\bottomrule[2.0pt]
		\end{tabular}
	}
	\caption{
		Results of different mapping networks with BLIP on FashionIQ validation set.
		The best results are shown in bold fonts.
	} 
	\label{tab:ablation mapping network fashionIQ}
\end{table}

Table \ref{tab:ablation light network fashionIQ} shows utilizing different query models for ZSCIR, which shows the superiority of lightweight models with pre-training initialization. 

\begin{table}[htb]
	\centering
	\renewcommand\arraystretch{1.1}
	\resizebox{0.92\textwidth}{0.11\textheight}{
		\begin{tabular}{ccccccccc}
			\toprule[2.0pt]
			\multirow{2}{*}{\textbf{Query model}} & \multicolumn{2}{c}{\textbf{Dress}} & \multicolumn{2}{c}{\textbf{Shirt}} &
			\multicolumn{2}{c}{\textbf{Top\&tee}} &
			\multicolumn{2}{c}{\textbf{Average}} \\
			\cline{2-9}
			& R@10 & R@50 & R@10 & R@50 & R@10 & R@50 & R@10 & R@50 \\ 
			\hline
			EfficientNet B0 & 25.19 & 45.56 & 28.95 & 48.04 & 32.53 & 52.78 & 28.89 & 48.79 \\ 
			EfficientNet B0* & 23.05 & 44.17 & 27.43 & 45.14 & 29.98 & 51.76 & 26.82 & 47.02 \\ 
			EfficientNet B2 & 25.33 & \textbf{46.26} & \textbf{30.03} & 48.58 & \textbf{33.45} & 53.80 & \textbf{29.60} & \textbf{49.54} \\ 
			EfficientNet B2* & 25.04 & 44.22 & 28.31 & 46.76 & 31.26 & 53.24 & 28.20 & 48.07 \\ 
			EfficientViT M2 & \textbf{25.48} & 45.51 & 29.64 & \textbf{48.68} & 32.94 & \textbf{54.31} & 29.35 & 49.50 \\ 
			EfficientViT M2* & 24.14 & 43.63 & 27.63 & 45.58 & 31.72 & 53.24 & 27.83 & 47.48 \\ 
			MobileNet V2 & 24.34 & 45.51 & 28.95 & 47.35 & 32.02 & 54.11 & 28.44 & 48.99 \\ 
			MobileNet V2* & 24.59 & 44.47 & 27.48 & 45.78 & 31.51 & 53.19 & 27.86 & 47.81 \\
			MobileViT V2 & 22.16 & 40.46 & 26.35 & 43.23 & 27.84 & 49.01 & 25.45 & 44.23 \\
			MobileViT V2* & 21.52 & 39.41 & 25.66 & 42.89 & 27.54 & 48.70 & 24.91 & 43.67 \\
			\bottomrule[2.0pt]
		\end{tabular}
	}
	\caption{
		Results of different query models with BLIP on FashionIQ validation set. 
		* denotes adopting random initialization for light query model.
		The best results are shown in bold fonts.
	} 
	\label{tab:ablation light network fashionIQ}
\end{table}

Table \ref{tab:ablation symmetrical fashioniq} shows the comparison of symmetric and asymmetric retrieval on FashionIQ dataset, and shows consistent superiority of our asymmetric structure.
\begin{table}[htb]
	\centering
	\renewcommand\arraystretch{1.1}
	\resizebox{0.88\textwidth}{0.065\textheight}{
		\begin{tabular}{cccccccccc}
			\toprule[2.0pt]
			\multirow{2}{*}{\textbf{Query model}} & \multirow{2}{*}{\textbf{Gallery model}} & \multicolumn{2}{c}{\textbf{Dress}} & \multicolumn{2}{c}{\textbf{Shirt}} & \multicolumn{2}{c}{\textbf{Toptee}} & \multicolumn{2}{c}{\textbf{Average}} \\
			\cline{3-10}
			& & R@10 & R@50 & R@10 & R@50 & R@10 & R@50 & R@10 & R@50 \\ 
			\hline
			EfficientNet B2 & BLIP VE & 25.33 & \textbf{46.26} & 30.03 & 48.58 & 33.45 & 53.80 & 29.60 & 49.54 \\ 
			EfficientNet B2 & BLIP VE & 0.5 & 2.08 & 0.64 & 2.01 & 0.87 & 1.63 & 0.67 & 1.91 \\ 
			EfficientViT M2 & BLIP VE & \textbf{25.48} & 45.51 & 29.64 & 48.68 & 32.94 & 54.31 & 29.35 & 49.50  \\
			EfficientViT M2 & BLIP VE & 0.69 & 1.59 & 1.08 & 2.16 & 1.22 & 2.04 & 1.00 & 1.93 \\ 
			BLIP VE $\dagger$ & BLIP VE & 24.89 & 45.36 & \textbf{30.18} & \textbf{49.07} & \textbf{35.65} & \textbf{55.12} & \textbf{30.24} & \textbf{49.85} \\
			BLIP VE & BLIP VE & 2.13 & 5.7 & 1.86 & 4.47 & 2.4 & 6.12 & 2.13 & 5.43 \\
			\bottomrule[2.0pt]
		\end{tabular}
	}
	\caption{
		Comparison with symmetric retrieval on the validation set of FashionIQ.
		BLIP VE denotes the BLIP visual encoder.
		$\dagger$ denotes updating the query BLIP VE while preserving the pre-trained BLIP VE in gallery side during training for maintaining the multi-modality alignment.
		The best results are shown in bold fonts..
	}
	\label{tab:ablation symmetrical fashioniq}
\end{table}

Table \ref{tab:ablation losses fashionIQ} demonstrates the influence of different losses on FashionIQ, which further verifies the effectiveness of GCD and LAR losses.

\begin{table}[htb]
	\centering
	\renewcommand\arraystretch{1.1}
	\resizebox{0.96\textwidth}{0.09\textheight}{
		\begin{tabular}{cccccccccc}
			\toprule[2.0pt]
			\multirow{2}{*}{\textbf{Query model}} & \multirow{2}{*}{\textbf{Variants}} & \multicolumn{2}{c}{\textbf{Dress}} & \multicolumn{2}{c}{\textbf{Shirt}} & \multicolumn{2}{c}{\textbf{Top\&tee}} & \multicolumn{2}{c}{\textbf{Average}} \\
			\cline{3-10}
			& & R@10 & R@50 & R@10 & R@50 & R@10 & R@50 & R@10 & R@50 \\ 
			\hline
			\multirow{4}{*}{EfficientNet B2} & GCD + LAR & \underline{25.33} & \textbf{46.26} & \textbf{30.03} & \underline{48.58} & \textbf{33.45} & \underline{53.80} & \textbf{29.60} & \textbf{49.54} \\ 
			& GCD & 23.90 & \underline{45.66} & 27.33 & 45.14 & 31.31 & 52.12 & 27.51 & 47.64 \\ 
			& LAR & 22.76 & 42.69 & 26.15 & 44.16 & 29.68 & 50.59 & 26.20 & 45.81 \\ 
			& GCD + LLM TR & 24.19 & 45.17 & 27.92 & 44.90 & 31.06 & 51.56 & 27.72 & 47.21 \\ 
			\midrule[1.5pt]
			\multirow{4}{*}{EfficientViT M2} & GCD + LAR & \textbf{25.48} & 45.51 & \underline{29.64} & \textbf{48.68} & \underline{32.94} & \textbf{54.31} & \underline{29.35} & \underline{49.50}  \\ 
			& GCD & 23.35 & 44.03 & 26.79 & 44.80 & 30.65 & 50.99 & 26.93 & 46.61 \\ 
			& LAR & 13.73 & 28.46 & 21.49 & 38.13 & 24.68 & 43.50 & 19.97 & 36.70 \\ 
			& GCD + LLM TR & 23.80 & 44.03 & 26.94 & 45.00 & 30.49 & 50.94 & 27.08 & 46.66 \\
			\bottomrule[2.0pt]
		\end{tabular}
	}
	\caption{Results of different losses on FashionIQ validation set. GCD: global contrastive distillation; LAR: local alignment regularization; LLM TR: text regularization that adopts the sentences made by LLM.
	The best and second-best results are highlighted in bold and underlined fonts, respectively.} 
	\label{tab:ablation losses fashionIQ}
\end{table}

To better understand the performance of zero-shot methods, we compare our methods with the supervised methods that adopt the labeled-triplets for training, including CIRPLANT \cite{liu2021image}, ARTEMIS \cite{delmas2021artemis}, and CLIP4Cir \cite{baldrati2022effective}.
We report the results on CIRR and fashionIQ, Since CIRCO does not include training set so that there is no supervised results for it.
From Table \ref{tab:supervised CIRR} and \ref{tab:supervised fashionIQ}, our method could surpass CIRPLANT and ALTEMIS, which shows the effectiveness of leveraging pre-trained BLIP model for zero-shot learning.
Although there is still gap compared with CLIP4Cir, our method is getting closer with the supervised state-of-the-art method in contrast to previous zero-shot methods, while achieving deployment flexibility with the asymmetric framework.   

\begin{table}[htb]
	\centering
	\renewcommand\arraystretch{1.1}
	\resizebox{0.76\textwidth}{0.1\textheight}{
		\begin{tabular}{ccccccc}
			\toprule[2.0pt]
			\multirow{2}{*}{\textbf{Supervision}} & \multirow{2}{*}{\textbf{Method}} & \multicolumn{4}{c}{\textbf{Recall@K}} & \multirow{2}{*}{\textbf{Avg}} \\
			\cline{3-6}
			& & R@1 & R@5 & R@10 & R@50 &  \\ 
			\hline
			\multirow{3}{*}{Supervised} & CIRPLANT & 19.60 & 52.60 & 68.40 & 92.40 & 58.25 \\
			& ALTEMIS & 16.96 & 46.10 & 61.31 & 87.73 & 53.03 \\
			& CLIP4Cir & \textbf{33.60} & \textbf{65.40} & \textbf{77.40} & \textbf{95.20} & \textbf{67.90} \\
			\midrule[1.5pt]
			\multirow{6}{*}{Zero-shot} & Pic2Word(CLIP) & 23.90 & 51.70 & 65.30 & 87.80 & 57.17 \\
			& SEARLE(CLIP) & 24.22 & 52.41 & 66.29 & 88.63 & 57.88 \\
			& Pic2word(BLIP) & 26.70 & 53.16 & 64.10 & 84.36 & 57.08 \\
			& SEARLE(BLIP) & 29.27 & 54.86 & 66.57 & 86.16 & 59.21 \\
			& Ours(Asy, EN B2) & \underline{30.84} & \underline{61.06} & \underline{73.57} & \underline{92.43} & \underline{64.48} \\ 
			& Ours(Asy, EVT M2) & 29.63 & 58.99 & 71.37 & 91.47 & 62.87 \\ 
			\bottomrule[2.0pt]
		\end{tabular}
	}
	\caption{
		Comparsion with supervised methods on CIRR test set.
		The best and second best results are shown in bold and underline fonts, respectively.
	} 
	\label{tab:supervised CIRR}
\end{table}

\begin{table}[htb]
	\centering
	\renewcommand\arraystretch{1.1}
	\resizebox{0.98\textwidth}{0.1\textheight}{
		\begin{tabular}{cccccccccc}
			\toprule[2.0pt]
			\multirow{2}{*}{\textbf{Supervision}} & \multirow{2}{*}{\textbf{Method}} & \multicolumn{2}{c}{\textbf{Dress}} & \multicolumn{2}{c}{\textbf{Shirt}} &
			\multicolumn{2}{c}{\textbf{Top\&tee}} &
			\multicolumn{2}{c}{\textbf{Average}} \\
			\cline{3-10}
			& & R@10 & R@50 & R@10 & R@50 & R@10 & R@50 & R@10 & R@50 \\ 
			\hline
			\multirow{3}{*}{Supervised} & CIRPLANT & 17.45 & 40.41 & 17.53 & 38.81 & 21.64 & 45.38 & 18.87 & 41.53 \\
			& ALTEMIS & 27.20 & 52.40 & 21.80 & 43.60 & 29.20 & 54.80 & 26.10 & 50.30 \\
			& CLIP4Cir & \textbf{30.30} & \textbf{54.50} & \textbf{37.20} & \textbf{55.80} & \textbf{39.20} & \textbf{61.30} & \textbf{35.60} & \textbf{57.20} \\
			\midrule[1.5pt]
			\multirow{6}{*}{Zero-shot} & Pic2Word(CLIP) & 20.00 & 40.20 & 26.20 & 43.60 & 27.90 & 47.40 & 24.70 & 43.70 \\
			& SEARLE(CLIP) & 20.32 & 43.18 & 27.43 & 45.68 & 29.32 & 50.17 & 25.69 & 46.34 \\
			& Pic2word(BLIP) & 21.35 & 42.68 & 27.51 & 46.01 & 29.12 & 49.33 & 25.99 & 46.00 \\
			& SEARLE(BLIP) & 22.11 & 41.79 & 29.72 & 48.53 & 31.03 & 52.37 & 27.62 & 47.56 \\
			& Ours(Asy, EN B2) & 25.33 & \underline{46.26} & \underline{30.03} & 48.58 & \underline{33.45} & 53.80 & \underline{29.60} & \underline{49.54} \\ 
			& Ours(Asy, EVT M2) & \underline{25.48} & 45.51 & 29.64 & \underline{48.68} & 32.94 & \underline{54.31} & 29.35 & 49.50 \\  
			\bottomrule[2.0pt]
		\end{tabular}
	}
	\caption{
		Comparsion with supervised methods on FashionIQ validation set.
		The best and second best results are shown in bold and underline fonts, respectively.
	} 
	\label{tab:supervised fashionIQ}
\end{table}

\begin{figure*}[htb]
	\begin{center}
		\includegraphics[width=1.0\textwidth]{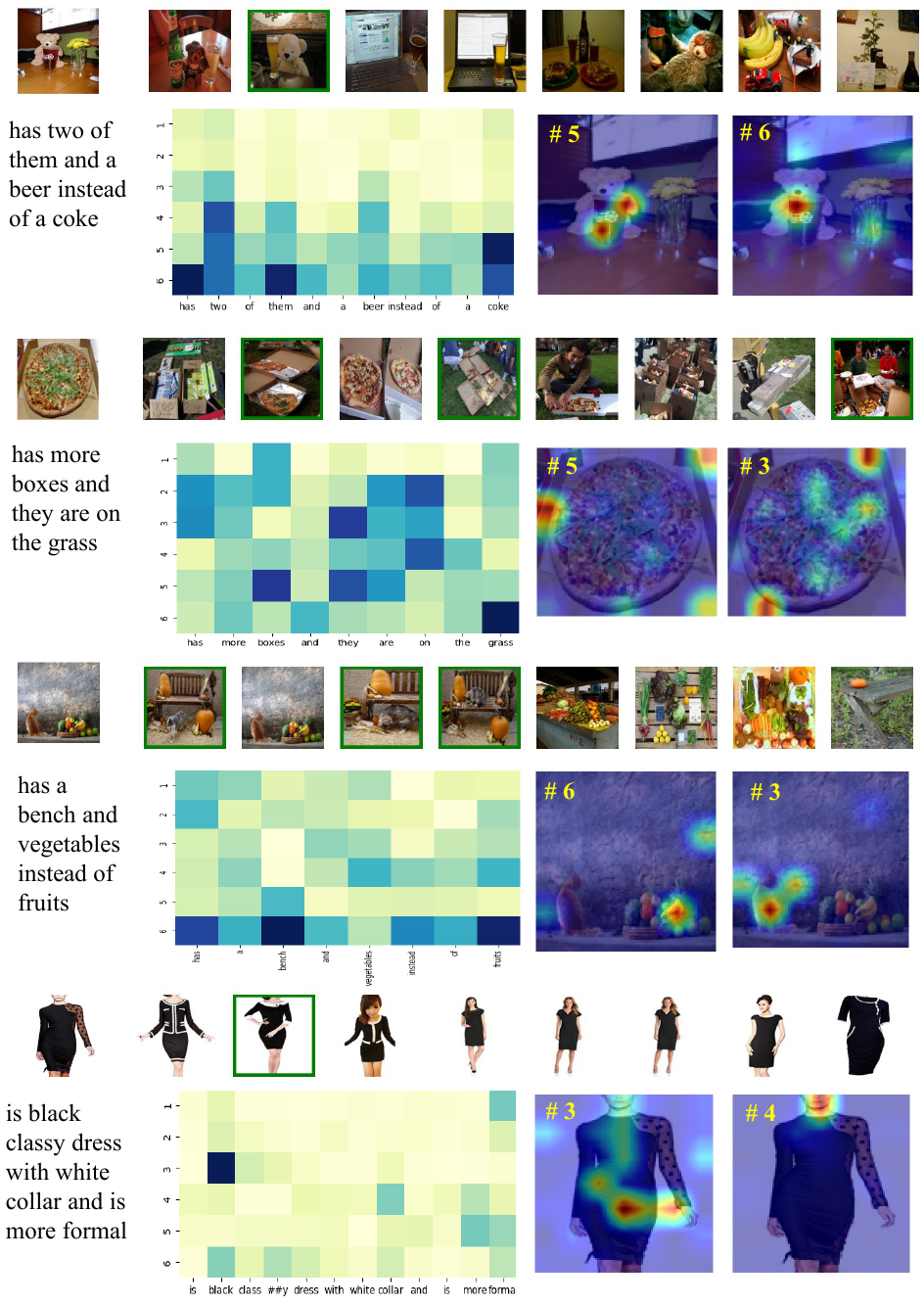} 
	\end{center}
	\caption{
		Some good retrieval examples and the attention map of sentence tokens on visual feature map and with modifier text. 
		Ground-truth target images are marked in green boxes.
	} 
	\label{retrieval good}
\end{figure*}

We visualize the results of the attention map of sentence tokens with the text modifier, and to compare them with the spatial attention of adaptive token learner together with the retrieval results.
Some good retrieval results are shown in Figure \ref{retrieval good}.
We could see that there are some tokens that attend to both concepts of query images and concepts of text modifier. 
In the up-right example, the third and the 6-th tokens focus on ``cat" and ``fruits" respectively, and the 4-th and 6-th are paying attention to the concepts in the text modifier, \textit{i.e.}, ``vegetables" and ``bench", respectively.
These examples could extract distinctive visual concepts and interact with the text modifier correspondingly, thus achieving good retrieval performance. 

\begin{figure*}[htb]
	\begin{center}
		\includegraphics[width=1.0\textwidth]{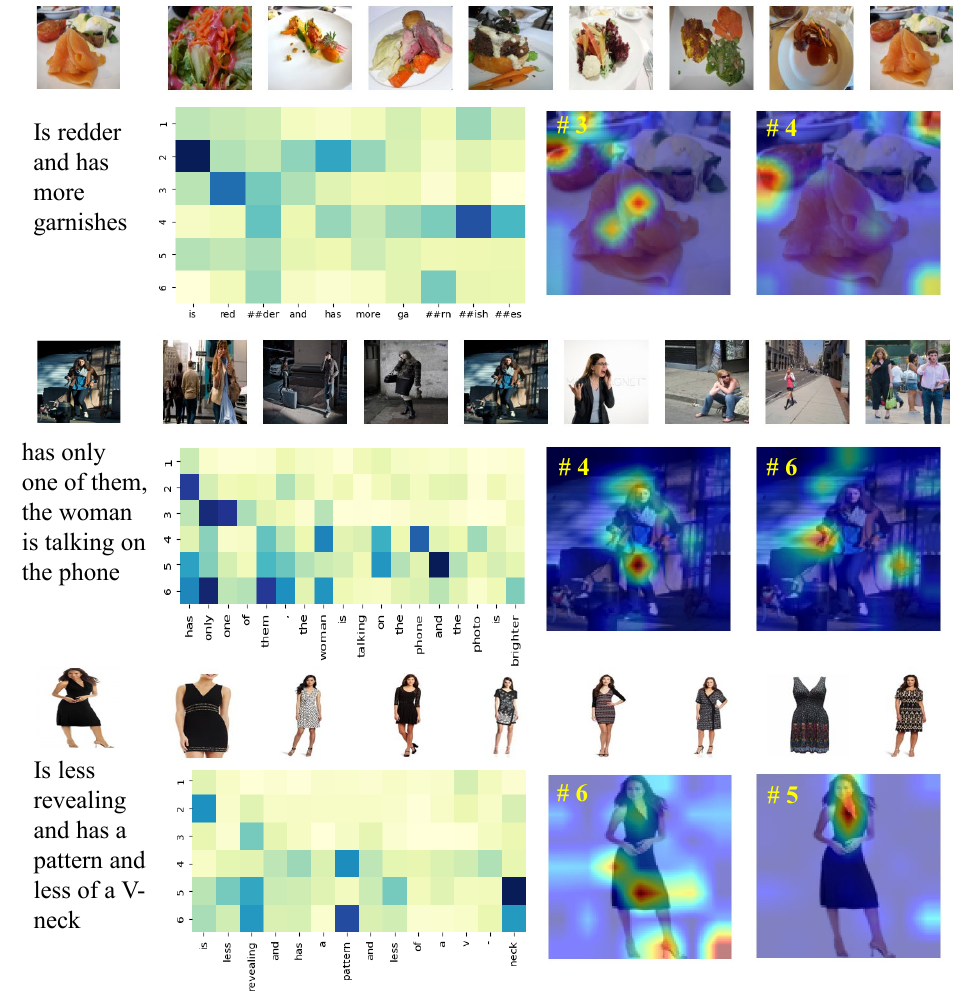} 
	\end{center}
	\caption{
		Some failure retrieval case and the attention map of sentence tokens on visual feature map and with modifier text. 
		Ground-truth target images are marked in green boxes.
	} 
	\label{retrieval bad}
\end{figure*}

We also present some failure cases in Figure \ref{retrieval bad}. 
The failure retrieval could be attributed to the wrong associated concept, rare words, and abstract concepts. 
For the example in the middle, although the 4-th token catches the concept of ``woman" and attends to the ``phone", there is difficulty associating ``them" in the text modifier with ``dog", \textit{i.e.}, the 6-th attends to ``them" but fail to catch the concept of ``dog" in the visual feature map.
For the example in the bottom, it is difficult for the model to grasp the abstract concept ``pattern" and ``less of a V-neck", thus leading to wrong retrieval results.  

\begin{figure*}[htb]
	\begin{center}
		\includegraphics[width=1.0\textwidth]{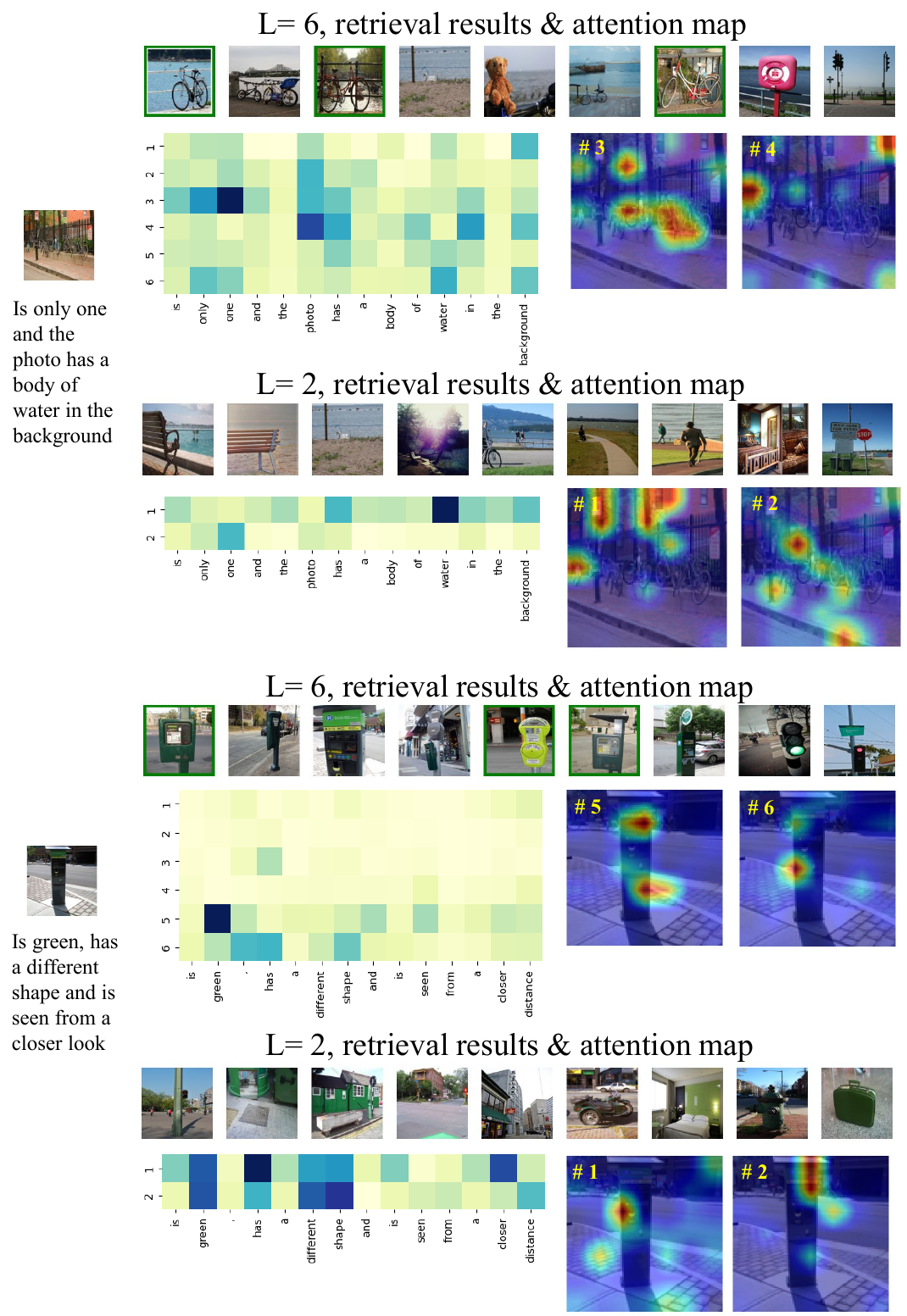} 
	\end{center}
	\caption{
		More comparison results of different token lengths (2 v.s. 6). 
		Query image and text modifier are shown on the left, and the attention map of sentence tokens with text modifiers and on visual feature map are shown on the right. 
		Ground-truth target images are marked in green boxes. 
	} 
	\label{length comparison 2 vs 6}
\end{figure*}

\begin{figure*}[htb]
	\begin{center}
		\includegraphics[width=0.95\textwidth]{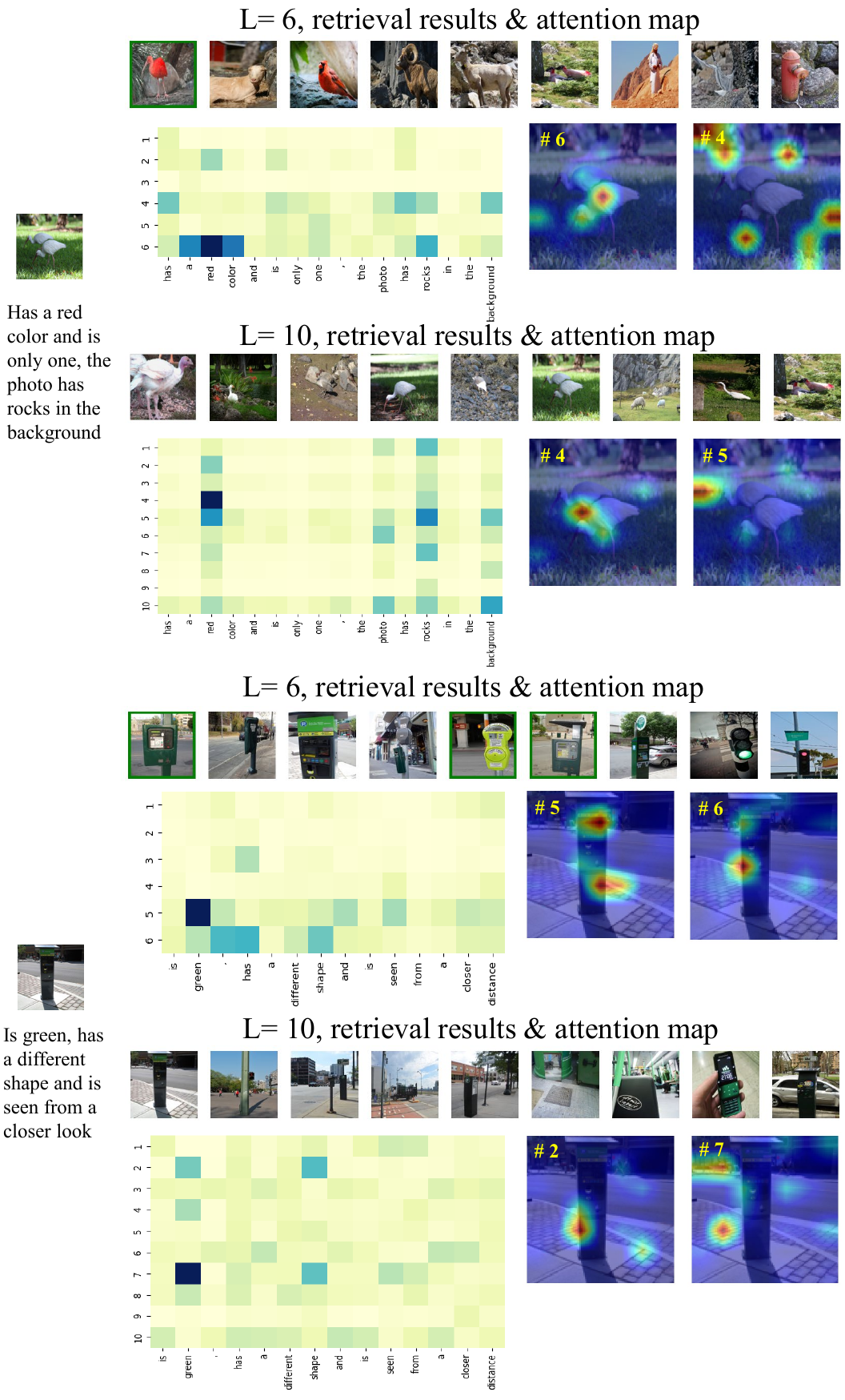} 
	\end{center}
	\caption{
		More comparison results of different token lengths (6 v.s. 10). 
		Query image and text modifier are shown on the left, and the attention map of sentence tokens with text modifiers and on visual feature map are shown on the right. 
		Ground-truth target images are marked in green boxes. 
	} 
	\label{length comparison 6 vs 10}
\end{figure*}

We present more visualization comparison of different token lengths in Figure \ref{length comparison 2 vs 6} and \ref{length comparison 6 vs 10}.
Generally, to achieve accurate retrieval, the pseudo sentence tokens need to accomplish two aspects: extracting effective visual information and interacting with textual information.
When the token length is small, there would be drop of performance due to the lack of capacity to extract sufficient visual information. As shown in Figure \ref{length comparison 2 vs 6}, query model with small token length performs worse even if the attention maps are reasonable.
When the token length increases for too long, on one hand, some tokens may focus on trivial and noisy visual patterns, as shown in Figure \ref{token length attention and retrieval}; on the other hand, these tokens may interact incorrectly with the text modifier, and may impact the correct token-text interaction.
As shown in Figure \ref{length comparison 6 vs 10}, the excessive tokens associate the visual pattern with incorrect text information, potentially interfering with the interaction of multi-modal information. 
Even if there are accurate associations between other sentence tokens and text modifier, this interference would still degrade the retrieval accuracy.

To verify the efficiency of our framework,  we report the latency on both query side and cloud side on CIRCO vailidation set as shown in Table \ref{tab:latency}. 
For query side, we report the inference latency for different query models. 
We find that lightweight encoders are generally twice faster than large visual encoder (BLIP VE). For cloud side, since the gallery models are the same (BLIP TE \& VE), the retrieval latency is the same.

To further verify the effectiveness of adaptive token learner, we carry out experiments for other three variants of mapping networks:

(I) simple tokenizer (ST): only using the block described in Eq. (\ref{spatial attention map}) and Eq. (\ref{visual token}) in section 3.1 in the manuscript.

(II) adaptive token learner without cross-attention (ATL w/o CA): including the blocks described in Eq. (\ref{spatial attention map}), (\ref{visual token}), (\ref{self-attention}).

(III) adaptive token learner without cross-attention (ATL w/o SA): including the blocks described in Eq. (\ref{spatial attention map}), (\ref{visual token}), (\ref{cross-attention}).

We report the results on EfficientNet B2 as below. From the table, it could be seen that with more components added to the mapping network, the performance gradually improves on three benchmark datasets. The simple tokenizer divides the semantics in different groups, which preserves more visual information than MLP. The self-attention models the mutual relationship between different visual tokens and cross-attention filters out noisy information, enhancing the pseudo sentence tokens to be more robust and refined visual representations in word embedding space. Therefore, the full version of adaptive token learner (ATL) achieves the best performance.

\begin{table}[htb]
	\centering
	\renewcommand\arraystretch{1.1}
	\resizebox{0.95\textwidth}{0.075\textheight}{
		\begin{tabular}{ccccc}
			\toprule[2.0pt]
			\textbf{Query model} & \textbf{\makecell[c]{Query inference \\ latency (ms, per query)}} & \textbf{Gallery model} & \textbf{\makecell[c]{Retrieval latency \\ (ms, per query)}} & \textbf{Total retrieval latency} \\
			\hline
			EfficientNet B0 & 3.07 & BLIP VE & 3.58 & 6.65 \\ 
			EfficientNet B2 & 3.24 & BLIP VE & 3.58 & 6.82 \\
			EfficientViT M2 & 3.27 & BLIP VE & 3.58 & 6.85 \\
			MobileNet V2 & 3.03 & BLIP VE & 3.58 & 6.61 \\
			MobileViT V2 & 3.40 & BLIP VE & 3.58 & 6.98 \\
			BLIP VE & 6.53 & BLIP VE & 3.58 & 10.11 \\
			CLIP ViT-L/14 & 24.44 & CLIP ViT-L/14 & 3.60 & 15.38 \\
			\bottomrule[2.0pt]
		\end{tabular}
	}
	\caption{
		Inference latency on query side and gallery side for different query and gallery models.
	} 
	\label{tab:latency}
\end{table}

\begin{table}[htb]
	\centering
	\renewcommand\arraystretch{1.1}
	\resizebox{0.76\textwidth}{0.065\textheight}{
		\begin{tabular}{ccccccccc}
			\toprule[2.0pt]
			\multirow{2}{*}{\textbf{Mapping network}} & \multicolumn{3}{c}{\textbf{Components}} & \multicolumn{4}{c}{\textbf{Recall@K}} & \multirow{2}{*}{\textbf{Avg}} \\
			\cline{2-8}
			& ST & SA & CA & R@1 & R@5 & R@10 & R@50 &  \\ 
			\hline
			MLP & & & & 30.18 & 59.46 & 71.30 & 89.33 & 62.57 \\
			ST & \checkmark & & & 29.94	& 61.68	& 73.71	& 91.92	& 64.31 \\
			ATL w/o CA & \checkmark & \checkmark &  & 29.85	& 61.61 & 74.48 & 91.53 & 64.37 \\
			ATL w/o SA & \checkmark & & \checkmark & 30.35 & 62.13 & 75.95 & 91.89 & 65.08 \\
			ATL & \checkmark & \checkmark & \checkmark & \textbf{31.48} & \textbf{63.42} & \textbf{77.27} & \textbf{93.16} & \textbf{66.33} \\
			\bottomrule[2.0pt]
		\end{tabular}
	}
	\caption{
		Ablation study on variants of mapping networks on CIRR validation set. Best results are shown in bold fonts.
	} 
	\label{tab:ablation ATL cirr}
\end{table}

\begin{table}[htb]
	\centering
	\renewcommand\arraystretch{1.1}
	\resizebox{0.76\textwidth}{0.06\textheight}{
		\begin{tabular}{ccccccccc}
			\toprule[2.0pt]
			\multirow{2}{*}{\textbf{Mapping network}} & \multicolumn{3}{c}{\textbf{Components}} & \multicolumn{4}{c}{\textbf{mAP@K}} & \multirow{2}{*}{\textbf{Avg}} \\
			\cline{2-8}
			& ST & SA & CA & mAP@1 & mAP@5 & mAP@10 & mAP@50 &  \\ 
			\hline
			MLP & & & & 8.28 & 8.67	& 9.62 & 10.00 & 9.14 \\
			ST & \checkmark & & & 9.89 & 10.33 & 11.38 & 11.79 & 10.85 \\
			ATL w/o CA & \checkmark & \checkmark &  & 10.46	& 10.94	& 12.03 &	12.52 & 11.49 \\
			ATL w/o SA & \checkmark & & \checkmark & 11.52 & 12.15 & 13.83 & 13.58 & 12.77 \\
			ATL & \checkmark & \checkmark & \checkmark & \textbf{13.19} & \textbf{13.83} & \textbf{15.20} & \textbf{15.85} & \textbf{14.52} \\
			\bottomrule[2.0pt]
		\end{tabular}
	}
	\caption{
		Ablation study on variants of mapping networks on CIRCO validation set. Best results are shown in bold fonts.
	} 
	\label{tab:ablation ATL circo}
\end{table}

\begin{table}[htb]
	\centering
	\renewcommand\arraystretch{1.1}
	\resizebox{0.98\textwidth}{0.06\textheight}{
		\begin{tabular}{cccccccccccc}
			\toprule[2.0pt]
			\multirow{2}{*}{\textbf{Mapping network}} & \multicolumn{3}{c}{\textbf{Components}} & \multicolumn{2}{c}{\textbf{Dress}} & \multicolumn{2}{c}{\textbf{Shirt}} &
			\multicolumn{2}{c}{\textbf{Top\&tee}} &
			\multicolumn{2}{c}{\textbf{Average}} \\
			\cline{2-12}
			& ST & SA & CA & R@10 & R@50 & R@10 & R@50 & R@10 & R@50 & R@10 & R@50 \\ 
			\hline
			MLP & & & & 23.65 & 43.83 & 28.46 & 47.74 & 31.16 & 51.96 & 27.76 & 47.84 \\
			ST & \checkmark & & & 23.45	& 43.58	& 28.80	& 45.78	& 31.57	& 52.32 & 27.94 & 47.23 \\
			ATL w/o CA & \checkmark & \checkmark & & 24.00 & 45.86 & 27.33 & 45.29 & 31.62 & 52.27 & 27.65 & 47.80 \\
			ATL w/o SA & \checkmark & & \checkmark & 24.15 & \textbf{46.71} & 28.16 & 46.22 & 32.46 & 52.37 & 28.26 & 48.43 \\
			ATL & \checkmark & \checkmark & \checkmark & \textbf{25.33} & 46.26 & \textbf{30.03} & \textbf{48.58} & \textbf{33.45} & \textbf{53.80} & \textbf{29.60} & \textbf{49.54} \\
			\bottomrule[2.0pt]
		\end{tabular}
	}
	\caption{
		Ablation study on variants of mapping networks on FashionIQ. Best results are shown in bold fonts.
	} 
	\label{tab:ablation ATL fashioniq}
\end{table}

\end{document}